\documentclass{article} 
\usepackage[table,dvipsnames]{xcolor}
\usepackage{iclr2025_conference,times}

\usepackage{mathtools}
\usepackage{hyperref}
\hypersetup{colorlinks=true,allcolors=[rgb]{0.2,0.2,0.75}}
\usepackage{url}
\usepackage{microtype}
\usepackage{subfigure}
\usepackage{wrapfig}
\usepackage{graphicx}
\usepackage{bbding}
\usepackage{amssymb}
\usepackage{amsmath}
\usepackage{amsfonts}
\usepackage{pifont}
\usepackage{booktabs}
\usepackage{multicol}
\usepackage{multirow}

\usepackage{listings}
\definecolor{codegreen}{rgb}{0,0.6,0}
\definecolor{codegray}{rgb}{0.5,0.5,0.5}
\definecolor{codepurple}{rgb}{0.58,0,0.82}
\definecolor{backcolour}{rgb}{0.95,0.95,0.92}
\lstdefinestyle{mystyle}{
    backgroundcolor=\color{backcolour},   
    commentstyle=\color{codegreen},
    keywordstyle=\color{magenta},
    numberstyle=\tiny\color{codegray},
    stringstyle=\color{codepurple},
    basicstyle=\ttfamily\scriptsize,
    breakatwhitespace=false,         
    breaklines=true,                 
    captionpos=b,                    
    keepspaces=true,                 
    numbers=left,                    
    numbersep=5pt,                  
    showspaces=false,                
    showstringspaces=false,
    showtabs=false,                  
    tabsize=2
}
\vspace{-0.3cm}
\title{EnzymeFlow: Generating Reaction-specific Enzyme Catalytic Pockets through Flow Matching and Co-Evolutionary Dynamics}
\vspace{-0.5cm}

\author{%
  Chenqing Hua$^{1,3}$ \ \ \ \ \ \ \ \  Yong Liu$^{5}$ \ \ \ \ \ \ \ \ Dinghuai Zhang$^{3,4}$ \ \ \ \ \ \ \ \ Odin Zhang$^{6}$ \ \ \ \ \ \ \ \ \vspace{-0.2cm} \And Sitao Luan $^{3}$ \ \ \ \ \ Kevin K. Yang $^{7}$ \ \ \ \ \ Guy Wolf $^{3,4}$ \ \ \ \ \ Doina Precup $^{1,3,8}$ \ \ \ \ \ Shuangjia Zheng$^{2}$\thanks{Correspondence to: \texttt{chenqing.hua@mail.mcgill.ca}; \ \ \texttt{shuangjia.zheng@sjtu.edu.cn}}
    \\
    $^1$McGill; \ \ $^2$SJTU; \ \ $^3$Mila-Quebec AI Institute; \ \ $^4$UdeM; \ \ $^5$HKUST; \\$^6$Institute for Protein Design, UW; \ \ $^7$Microsoft Research; \ \ $^8$DeepMind  \\
}

\iclrfinalcopy 
\begin{document}

\maketitle

\vspace{-0.5cm}
\begin{abstract}
\vspace{-0.3cm}
Enzyme design is a critical area in biotechnology, with applications ranging from drug development to synthetic biology. Traditional methods for enzyme function prediction or protein binding pocket design often fall short in capturing the dynamic and complex nature of enzyme-substrate interactions, particularly in catalytic processes. To address the challenges, we introduce EnzymeFlow, a generative model that employs flow matching with hierarchical pre-training and enzyme-reaction co-evolution to generate catalytic pockets for specific substrates and catalytic reactions. Additionally, we introduce a large-scale, curated, and validated dataset of enzyme-reaction pairs, specifically designed for the catalytic pocket generation task, comprising a total of $328,192$ pairs. By incorporating evolutionary dynamics and reaction-specific adaptations, EnzymeFlow becomes a powerful model for designing enzyme pockets, which is capable of catalyzing a wide range of biochemical reactions. Experiments on the new dataset demonstrate the model's effectiveness in designing high-quality, functional enzyme catalytic pockets, paving the way for advancements in enzyme engineering and synthetic biology. We provide EnzymeFlow code at \url{https://github.com/WillHua127/EnzymeFlow} with notebook demonstration at \url{https://github.com/WillHua127/EnzymeFlow/blob/main/enzymeflow_demo.ipynb}.
\end{abstract}

\vspace{-0.5cm}
\section{Introduction}
\vspace{-0.2cm}
Proteins are fundamental to life, participating in many essential interactions for biological processes \citep{whitford2013proteins}. 
\begin{wrapfigure}{R}{0.4\textwidth}
\vspace{-1cm}
  \centering
    \includegraphics[width=0.4\textwidth]{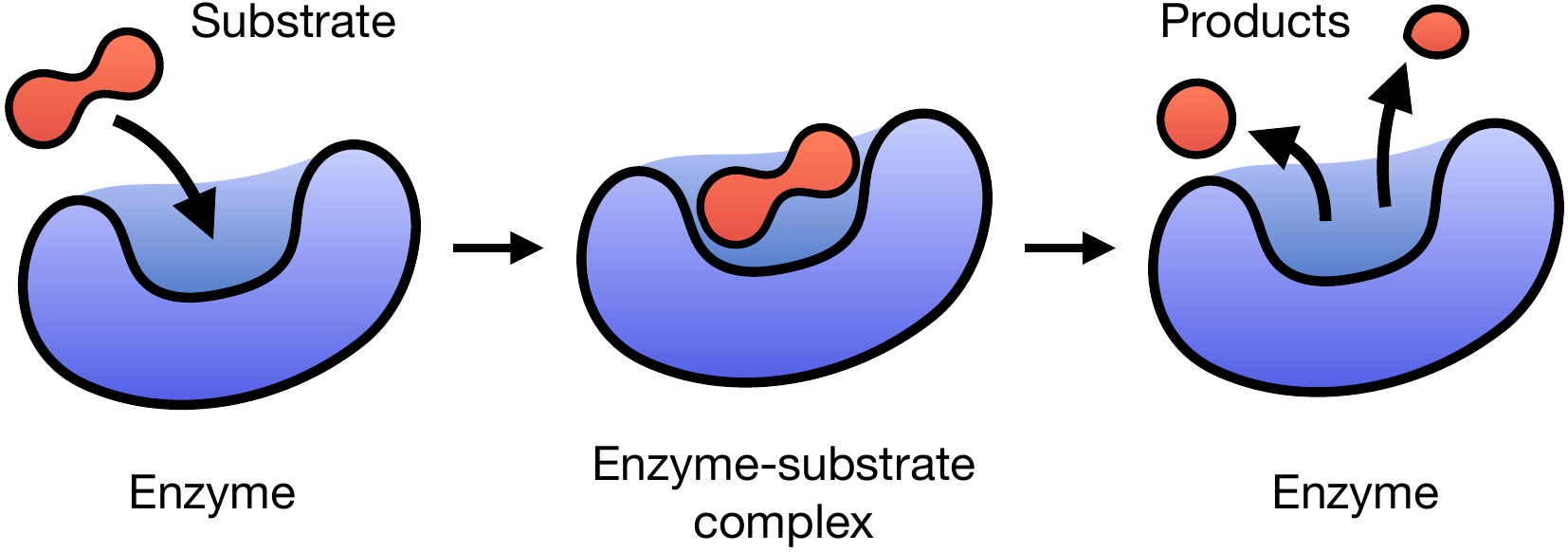}
  \vspace{-0.8cm}
  \caption{Enzyme-substrate Mechanism.
  \vspace{-0.6cm}
  }
  \label{fig:intro}
\end{wrapfigure}
Among proteins, enzymes stand out as a specialized class that serves as catalysts, driving and regulating nearly all chemical reactions and metabolic pathways across living organisms, from simple bacteria to complex mammals  \citep{kraut1988enzymes, murakami1996artificial, copeland2023enzymes} (visualized in Fig.~\ref{fig:intro}). 
Their catalytic power is central to biological functions, enabling the efficient production of complex organic molecules in biosynthesis \citep{ferrer2008structure, liu2007cofactor} and the creation of novel biological pathways in synthetic biology \citep{girvan2016applications, keasling2010manufacturing, hodgman2012cell}. Examining enzyme functions across the tree of life deepens our understanding of the evolutionary processes that shape metabolic networks and enable organisms to adapt to their environments \citep{jensen1976enzyme, glasner2006evolution, campbell2016role, pinto2022exploiting}. Consequently, studying enzyme-substrate interactions is essential for comprehending biological processes and designing effective products.

Traditional methods have primarily focused on enzyme function prediction, annotation \citep{gligorijevic2021structure, yu2023enzyme}, or enzyme-reaction retrieval \citep{mikhael2024clipzyme, hua2024reactzyme, yang2024care}. These approaches lack the ability to design new enzymes that catalyze specific biological processes. Recent studies suggest that current function prediction models struggle to generalize to unseen enzyme reaction data \citep{de2024limitations, kroll2023general}, limiting their utility in enzyme design. To effectively design enzymes, it is crucial not only to predict protein functions but also to identify and generate enzyme catalytic pockets specific to particular substrates, thereby enabling potentially valuable biological processes.

On the other hand, recent advances in deep generative models have significantly improved pocket design for protein-ligand complexes \citep{stark2023harmonic, zhang2023full, zhang2024pocketgen, krishna2024generalized}, generating diverse and functional binding pockets for ligand molecules. 
However, these models cannot generalize directly to the design of enzyme catalytic pockets for substrates involved in catalytic processes. Unlike protein-ligand complexes, where ligand binding typically does not lead to a chemical transformation, \textbf{enzyme-substrate interactions result in a chemical change where the substrate is converted into a product,} which has significantly different underlying mechanisms. More specifically, in protein-ligand binding, the ligand may induce a conformational change in the protein, affect its interactions with other molecules, or modulate its activity; in contrast, the formation of an enzyme-substrate complex is a precursor to a catalytic reaction, where the enzyme lowers the activation energy, facilitating the transformation of the substrate into a product. After the reaction, the enzyme is free to bind another substrate molecule. Therefore, current generative models for pocket design are restricted and limited to static ligand-binding interactions, failing to describe such dynamic transformations and the complex nature of enzyme-substrate interactions.

To address these limitations, we propose EnzymeFlow (demonstrated in Fig.~\ref{fig:enzymeflow}), a flow matching model \citep{lipman2022flow,liu2022flowstraightfastlearning,albergo2023buildingnormalizingflowsstochastic} 
with enzyme-reaction co-evolution and structure-based pre-training for enzyme catalytic pocket generation.  Our major contributions follow: \textbf{(1) EnzymeFlow—Flow Model for Enzyme Catalytic Pocket Design:} We define conditional flows for enzyme catalytic pocket generation based on backbone frames, amino acid types, and Enzyme Commission (EC) class. The generative flow process is conditioned on specific substrates and products, enabling potential catalytic processes. 
\textbf{(2) Enzyme-Reaction Co-Evolution:} Since enzyme-substrate interactions involve dynamic chemical transformations of substrate molecules, which is distinct from static protein-ligand interactions, we propose enzyme-reaction co-evolution with a new co-evolutionary transformer (\textit{coEvoFormer}). The co-evolution is used to capture substrate-specificity in catalytic reactions. It encodes how enzymes and reactions evolve together, allowing the model to operate on evolutionary dynamics, which naturally comprehends the catalytic process. 
\textbf{(3) Structure-Based Hierarchical Pre-Training:} To leverage the vast data of geometric structures from existing proteins and protein-ligand complexes, we propose a structure-based hierarchical pre-training. This method progressively learns from protein backbones to protein binding pockets, and finally to enzyme catalytic pockets. This hierarchical learning of protein structures enhances geometric awareness within the model. 
\textbf{(4) EnzymeFill—Large-scale Pocket-specific Enzyme-Reaction Dataset with Pocket Structures:} Current enzyme-reaction datasets are based on full enzyme sequences or structures and lack precise geometry for how enzyme pockets catalyze the substrates. To address this, we construct a structure-based, curated, and validated enzyme catalytic pocket-substrate dataset, specifically designed for the catalytic pocket generation task.

\vspace{-0.2cm}
\section{Related Work}

\vspace{-0.2cm}
\subsection{Protein Evolution}
\label{sec:2.1}
\vspace{-0.2cm}
Protein evolution learns how proteins change over time through processes such as mutation, selection, and genetic drift \citep{pal2006integrated, bloom2009light}, which influence protein functions. Studies on protein evolution focus on understanding the molecular mechanisms driving changes in protein sequences and structures. \cite{zuckerkandl1965molecules} introduce the concept of the molecular clock, which postulates that proteins evolve at a relatively constant rate over time, providing a framework for estimating divergence times between species. \cite{depristo2005missense} show that evolutionary rates are influenced by functional constraints, with regions critical to protein function (\textit{e.g.}, active sites, binding interfaces) evolving more slowly due to purifying selection. This understanding leads to the development of methods for detecting functionally important residues based on evolutionary conservation.
Understanding protein evolution has practical applications in protein engineering. By studying how natural proteins evolve to acquire new functions, researchers design synthetic proteins with desired properties \citep{xia2004simulating, jackel2008protein}. Additionally, deep learning models increasingly integrate evolutionary principles to predict protein function and stability, design novel enzymes, and guide protein engineering \citep{yang2019machine, alquraishi2019end, jumper2021highly}.

\vspace{-0.2cm}
\subsection{Generative Models for Protein and Pocket Design}
\label{sec:2.2}
\vspace{-0.2cm}
Recent advancements in generative models have advanced the field of protein design and binding pocket design, enabling the creation of proteins or binding pockets with desired properties and functions \citep{yim2023fast, yim2023se, chu2024all, hua2024effective, abramson2024accurate}. For example, RFDiffusion \citep{watson2023novo} employs denoising diffusion in conjunction with RoseTTAFold \citep{baek2021accurate} for \textit{de novo} protein structure design, achieving wet-lab-level generated structures that can be extended to binding pocket design. RFDiffusionAA \citep{krishna2024generalized} extends RFDiffusion for joint modeling of protein and ligand structures, generating ligand-binding proteins and further leveraging MPNNs for sequence design. Additionally, FAIR \citep{zhang2023full} and PocketGen \citep{zhang2024pocketgen} use a two-stage coarse-to-fine refinement approach to co-design pocket structures and sequences. Recent models leveraging flow matching frameworks have shown promising results in these tasks. For instance, FoldFlow \citep{bose2023se} introduces a series of flow models for protein backbone design, improving training stability and efficiency. FrameFlow \citep{yim2023fast} further enhances sampling efficiency and demonstrates success in motif-scaffolding tasks using flow matching, while MultiFlow \citep{campbell2024generative} advances to structure and sequence co-design. These flow models, initially applied to protein backbones, have been further generalized to binding pockets. For example, PocketFlow \citep{generalized2024zhang} combines flow matching with physical priors to explicitly learn protein-ligand interactions in binding pocket design, achieving stronger results compared to RFDiffusionAA.
While these models excel in protein and binding pocket design, they primarily focus on static protein(-ligand) interactions and lack the ability to model the chemical transformations involved in enzyme-catalyzed reactions. This limitation may reduce their accuracy and generalizability in designing enzyme pockets for catalytic reactions. In EnzymeFlow, we aim to address these current limitations. An extended discussion of related works on AI-driven protein engineering can be found in App.~\ref{app:related.work}.

\textbf{Discussion regarding PocketFlow.}
PocketFlow \citep{generalized2024zhang} has demonstrated strong performance in protein-ligand design, showing generalizability across various protein pocket categories. However, it falls short when applied to the design of enzyme catalytic pocket with specific substrates. One key limitation is that protein-ligand interactions are static, meaning that the training data and model design do not capture or describe the chemical transformations, such as the conversion or production of new molecules, that occur during enzyme-catalyzed reactions. This dynamic aspect of enzyme-substrate interactions is missing in current models. Another limitation is that PocketFlow fixes the overall protein backbone structure before designing the binding pocket, treating the pocket as a missing element to be filled in. This approach may not align with practical needs, as the overall protein backbone structure is often unknown before pocket design. Ideally, the design process should be reversed: the pocket should be designed first, influencing the overall protein structure.
Despite these challenges, PocketFlow remains a good and leading work in pocket design. With EnzymeFlow, we aim to address these limitations, particularly in the context of catalytic pocket design.

\vspace{-0.2cm}
\section{EnzymeFill: Large-scale Enzyme Pocket-Reaction Dataset}
\vspace{-0.3cm}
\label{sec:enzymeflow.dataset}

A key limitation of current datasets, such as ESP \citep{kroll2023turnover}, EnzymeMap \citep{heid2023enzymemap}, CARE \citep{yang2024care}, or ReactZyme \citep{hua2024reactzyme}, is the lack of precise pocket information. These datasets typically provide enzyme-reaction data, including protein sequences and SMILES representations, which is used to predict EC numbers in practice. To address it, we introduce a new synthetic dataset, EnzymeFill, which includes precise pocket structures with substrate conformations. EnzymeFill is specifically introduced for enzyme catalytic pocket design. 

\textbf{Data Source.}
We construct a curated and validated dataset of enzyme-reaction pairs by collecting data from the Rhea \citep{bansal2022rhea}, MetaCyc \citep{caspi2020metacyc}, and Brenda \citep{schomburg2002brenda} databases. For enzymes in these databases, we exclude entries missing UniProt IDs or protein sequences. For reactions, we apply the following procedures: (1) remove cofactors, small ion groups, and molecules that appear in both substrates and products within a single reaction; (2) exclude reactions with more than five substrates or products; and (3) apply OpenBabel \citep{o2011open} to standardize canonical SMILES. Ultimately, we obatin a total of $328,192$ enzyme-reaction pairs, comprising $145,782$ unique enzymes and $17,868$ unique reactions; we name it EnzymeFill.

\begin{figure*}[t!]
\vspace{-0.6cm}
\centering
{
\includegraphics[width=0.95\textwidth]{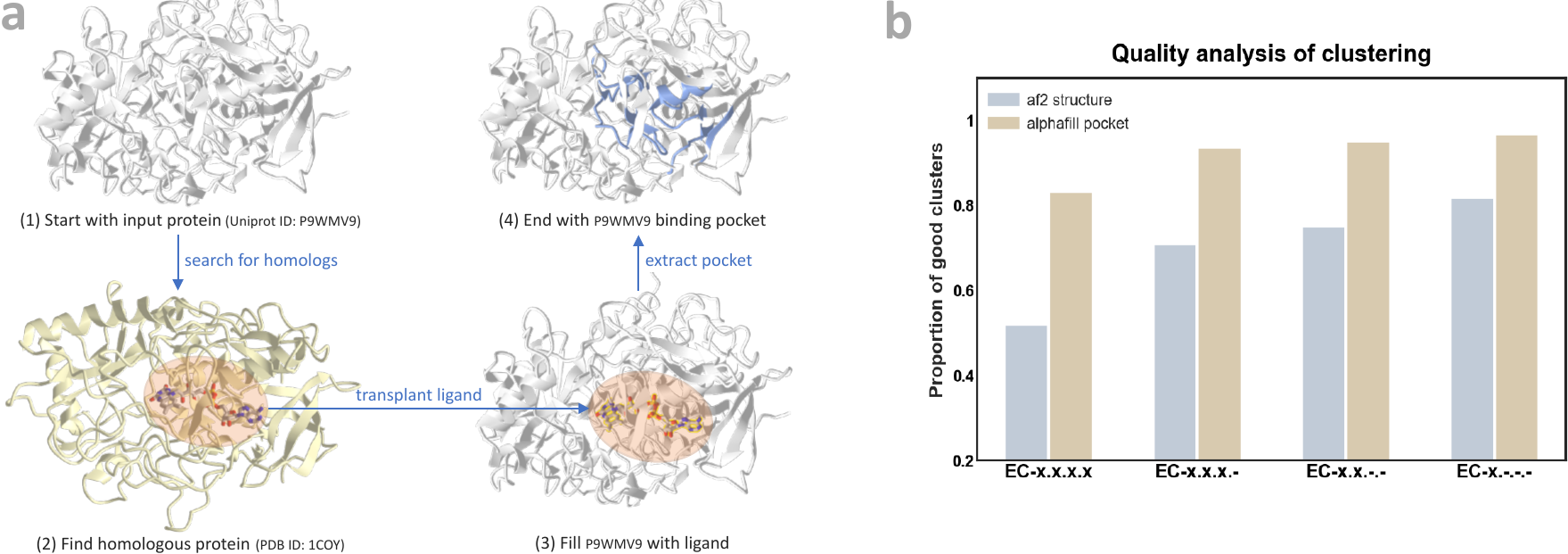}}
{
\vspace{-0.3cm}
  \caption{\textbf{(a)} Enzyme pocket extraction workflow with AlphaFill. \textbf{(b)} Quality analysis of clustering between enzyme pockets and full structures; good clusters have high functional concentration.}
  \label{fig:pocket.extraction}
  \vspace{-0.5cm}
}
\end{figure*}

\textbf{Catalytic Pocket with AlphaFill.}
We identify all enzyme catalytic pockets using AlphaFill \citep{hekkelman2023alphafill}, an AF-based algorithm that uses sequence and structure similarity to transplant ligand molecules from experimentally determined structures to predicted protein models. We download the AlphaFold structures for all enzymes and apply AlphaFill to extract the enzyme pockets. Simultaneously, we determine the reaction center by using atom-atom mapping of the reactions.
During the pocket extraction process, AlphaFill first identifies homologous proteins of the target enzyme in the PDB-REDO database, along with their complexes with ligands \citep{van2018homology}. It then transplants the ligands from the homologous protein complexes to the target enzyme through structural alignment (illustrated in Fig.~\ref{fig:pocket.extraction}(a)). After transplantation, we select the appropriate ligand molecule based on the number of atoms and its frequency of occurrence, and extract the pocket using a pre-defined radius of \small$10\mathring{\text{A}}$ \normalsize. We also perform clustering analysis on the extracted pockets using Foldseek \citep{van2022foldseek}, which reveals that enzyme catalytic pockets capture functional information more effectively than full structures (illustrated in Fig.~\ref{fig:pocket.extraction}(b)).
For the extraction of reaction centers, we first apply RXNMapper to extract atom-atom mappings \citep{schwaller2021extraction}, which maps the atoms between the substrates and products. We then identify atoms where changes occurred in chemical bonds, charges, and chirality, labeling these atoms as reaction centers.

\textbf{Data Debiasing for Generation.}
To ensure the quality of catalytic pocket data for the design task, we exclude pockets with fewer than $32$ residues\footnote{$32$ residues are chosen based on LigandMPNN \citep{dauparas2023atomic}, ensuring high-quality interactions.}, resulting in $232,520$ enzyme-reaction pairs.
Additionally, enzymes and their catalytic pockets can exhibit significant sequence similarity. When enzymes that are highly similar in sequence appear too frequently in the dataset, they tend to belong to the same cluster or homologous group, which can introduce substantial biases during model training. To mitigate this issue and ensure a more balanced dataset, it is important to reduce the number of homologous enzymes by clustering and selectively removing enzymes from the same clusters. This helps to debias the data and improve the model's generalizability. We perform sequence alignment to cluster enzymes and identify homologous ones \citep{steinegger2017mmseqs2}.
We then revise the dataset into five major categories based on enzyme sequence similarity, resulting in: (1) $19,379$ pairs with at most $40\%$ homology, (2) $34,750$ pairs with at most $50\%$ homology, (3) $53,483$ pairs with at most $60\%$ homology, (4) $100,925$ pairs with at most $80\%$ homology, and (5) $132,047$ pairs with at most $90\%$ homology. In EnzymeFlow, we choose to use the clustered data with at most $60\%$ homology with $53,483$ samples for training.
We provide more dataset statistics in App.~\ref{app:data.stats}

\vspace{-0.2cm}
\section{EnzymeFlow}
\vspace{-0.2cm}

\begin{figure*}[t!]
\vspace{-0.5cm}
\centering
{
\includegraphics[width=1.\textwidth]{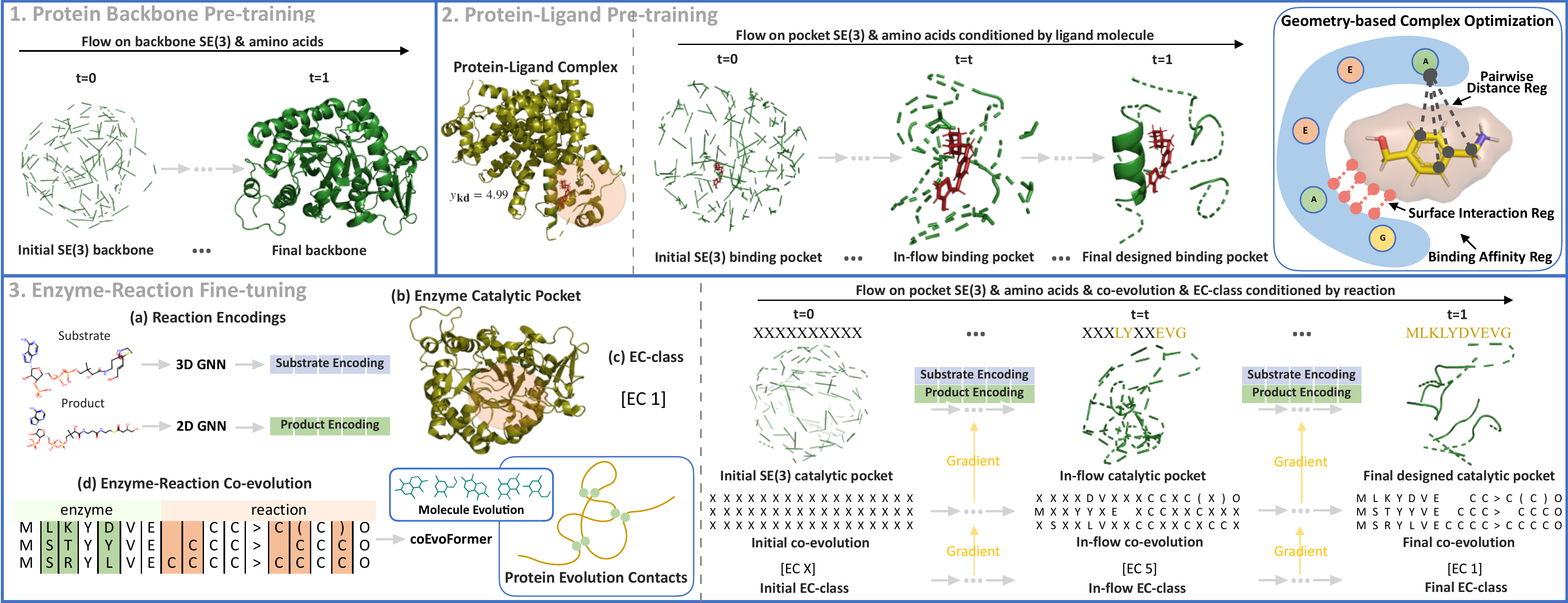}}
{
\vspace{-0.7cm}
  \caption{Overview of EnzymeFlow with hierarchical pre-training and enzyme-reaction co-evolution. \textbf{(1)} Flow model pre-trained on protein backbones and amino acid types. \textbf{(2)} Flow model further pre-trained on protein binding pockets, conditioned on ligand molecules with geometry-specific optimization. \textbf{(3)} Flow model fine-tuned on enzyme catalytic pockets, and conditioned on substrate and product molecules, with enzyme-reaction co-evolution and EC-class generation.}
  \label{fig:enzymeflow}
  \vspace{-0.5cm}
}
\end{figure*}

We introduce EnzymeFlow, a flow matching model with hierarchical pre-training and enzyme-reaction co-evolution to generate enzyme catalytic pockets for specific substrates and catalytic reactions. We demonstrate the pipeline in Fig.~\ref{fig:enzymeflow}, discuss the EnzymeFlow with co-evolutionary dynamics in Sec.~\ref{sec:general.enzymeflow}, further introduce the structure-based hierarchical pre-training for generalizability in Sec.~\ref{sec:pretraining}

\vspace{-0.2cm}
\subsection{Enzyme Catalytic Pocket Generation with Flow Matching}
\label{sec:general.enzymeflow}
\vspace{-0.2cm}

\textbf{EnzymeFlow on Catalytic Pocket.}
Following \cite{yim2023fast}, we refer to the protein structure as the backbone atomic coordinates of each residue. A pocket with number of residues \(N_r\) can be parameterized into SE(3) residue frames \(\{(x^i, r^i, c^i)\}_{i=1}^{N_r}\), where \(x^i \in \mathbb{R}^3\) represents the position (translation) of the \(C_\alpha\) atom of the \(i\)-th residue, \(r^i \in \text{SO(3)}\) is a rotation matrix defining the local frame relative to a global reference frame, and \small\(c^i \in \{1, \dots, 20\} \cup \{ \text{\ding{53}} \}\) \normalsize denotes the amino acid type, with an additional \text{\ding{53}} indicating a \textit{masking state} of the amino acid type. We refer to the residue block as \(T^i = (x^i, r^i, c^i)\), and the entire pocket is described by a set of residues \small\(\mathbf{T} = \{T^i\}_{i=1}^{N_r}\) \normalsize. Additionally, we denote the graph representations of substrate and product molecules in the catalytic reaction as \(l_s\) and \(l_p\), respectively. An enzyme-reaction pair can therefore be described as \((\mathbf{T}, l_s, l_p)\).

Following flow matching literature \citep{yim2023fast, campbell2024generative}, we use time $t=1$ to denote the source data.
The conditional flow on the enzyme catalytic pocket \(p_{t}(\mathbf{T}_t|\mathbf{T}_1)\) for a time step $t\in (0,1]$ can be factorized into the probability density over continuous variables (translations and rotations) and the probability mass function over discrete variables (amino acid types) as:
\vspace{-0.1cm}
\begin{equation}
\vspace{-0.1cm}
\label{eq:backbone.gen}
    p_{t}(\mathbf{T}_t|\mathbf{T}_1) = \prod_{i=1}^{N_r} p_{t}(x^i_t|x^i_1)\ p_{t}(r^i_t|r^i_1)\ p_{t}(c^i_t|c^i_1),
\end{equation}
where the translation, rotation, and amino acid type at time \(t\) are derived as:
\vspace{-0.1cm}
\small
\begin{equation}
\vspace{-0.1cm}
\begin{split}
x^i_t &= (1-t)x^i_0 + tx^i_1, \ x^i_0 \sim \mathcal{N}(0, I); \ \ r^i_t = \exp_{r^i_0}(t \log_{r^i_0} r^i_1), \ r^i_0 \sim \mathcal{U}_{\text{SO(3)}};  \\
c^i_t &\sim p_{t}(c^i_t|c^i_1) = \text{Cat}(t\ \delta (c^i_1, c^i_t) + (1-t)\ \delta (\text{\ding{53}}, c^i_t)),
\end{split}
\end{equation}
\normalsize
where \(\delta(\text{a}, \text{b})\) is the Kronecker delta, which equals to \(1\) if \(\text{a} = \text{b}\) and \(0\) if \(\text{a} \neq \text{b}\); Cat is a categorical distribution for the sampling of discrete amino acid type, with probabilities \small\(t \delta(c^i_1, c^i_t) + (1-t) \delta(\text{\ding{53}}, c^i_t)\)\normalsize. The discrete flow interpolates from the \textit{masking state} \text{\ding{53}} at \(t = 0\) to the actual amino acid type \(c^i_1\) at \(t = 1\) \citep{campbell2024generative}. In a catalytic process, enzymes interact with substrates to produce specific products. In practical enzyme design, we typically know the substrates $l_s$ (as 3D atom point clouds) and the desired products $l_p$ (as 2D molecular graphs or SMILES). Therefore, the formation of the enzyme catalytic pocket should be conditioned on both substrates and products. Our enzyme flow matching model is conditioned on these two ligand molecules $l_s, l_p$, ensuring that the predictions of vector fields ${v}_\theta(\cdot)$ and loss functions account for the substrate and product molecules:
\vspace{-0.1cm}
\small
\begin{equation}
\vspace{-0.1cm}
\label{eq:backbone.loss}
\begin{split}
    \mathcal{L}_{\text{trans}} &= \sum_{i=1}^{N_r} 	\| {v}_\theta^i(x^i_t, t, l_s, l_p)- (x^i_1 - x^i_0) \|^2_2; \ 
    \mathcal{L}_{\text{rot}} = \sum_{i=1}^{N_r} 	\| {v}_\theta^i(r^i_t, t, l_s, l_p)- \frac{\log_{r^i_t} r_1^i}{1-t} \|^2_\text{SO(3)}; \\
    \mathcal{L}_{\text{aa}} &= -\sum_{i=1}^{N_r} \log p_\theta(c^i_1 | {v}^i_\theta(c^i_t, t, l_s, l_p)).
\end{split}
\end{equation}
\normalsize
To design the enzyme pocket and model protein-ligand interactions, we implement 3D and 2D GNNs to encode the substrate and product, respectively (implemented in App.~\ref{app:mol.gnn}). The main vector field network applies cross-attention to model protein-ligand interactions and incorporates Invariant Point Attention (IPA) \citep{jumper2021highly} to encode protein features and make predictions. Following tricks in \cite{yim2023fast, campbell2024generative}, we let the the model predict the final structure at $t=1$ and interpolates to compute the vector fields (discussed in App.~\ref{app:vec.field}).

\textbf{EnzymeFlow on EC-Class.}
The Enzyme Commission (EC) classification is crucial for categorizing enzymes based on the reactions they catalyze. Understanding the EC-class of an enzyme-reaction pair can help predict its function in various biochemical pathways \citep{bansal2022rhea}. 
Given its importance, EnzymeFlow leverages EC-class to enhance its generalizability across various enzymes and catalytic reactions. Therefore, our model incorporates EC-class, \small$y_{\text{ec}} \in \{1,\dots,7\} \cup \{\text{\ding{53}}\}$ \normalsize, as a discrete factor in the design process. 
The EC-class is sampled from a Categorical distribution with probabilities \small\(t  \delta(y_{\text{ec}_1}, y_{\text{ec}_t}) + (1-t) \delta(\text{\ding{53}}, y_{\text{ec}_t})\) \normalsize. The discrete flow on EC-class interpolates from the \textit{masking state} \text{\ding{53}} at \(t=0\) to the actual EC-class \(y_{\text{ec}_1}\) at \(t=1\). The prediction and loss function are conditioned on the pocket frames and the substrate and product molecules:
\vspace{-0.1cm}
\begin{equation}
\vspace{-0.1cm}
    \mathcal{L}_\text{ec} = -\log p_\theta(y_{\text{ec}_1} | {v}_\theta(\mathbf{T}_t, t, l_s, l_p, y_{\text{ec}_t})).
\end{equation}
The model predicts the final EC-class at \(t=1\) and interpolates to compute its vector field. For EC-class prediction, we first employ a EC-class embedding network to encode \(y_{\text{ec}_t}\).  The final predicted EC-class is obtained by pooling cross-attention between the encoded enzyme and EC-class features.

\vspace{-0.2cm}
\subsubsection{EnzymeFlow with Enzyme-reaction co-Evolution}
\label{sec:coevo.enzymeflow}
\vspace{-0.2cm}

Enzyme (protein) evolution refers to the process by which enzyme structures and functions change over time due to genetic variations, such as mutations, duplications, and recombinations. These changes can lead to alterations in amino acids, potentially affecting the enzyme structure, function, stability, and interactions \citep{pal2006integrated, sikosek2014biophysics}.
Reaction evolution, on the other hand, refers to the process by which chemical reactions or substrates, particularly those catalyzed by enzymes, change and diversify within biological systems over time (illustrated in Fig.~\ref{fig:enzymeflow}(3)(d)).

\textbf{Co-Evolutionary Dynamics.} 
Enzymes can co-evolve with the metabolic or biochemical pathways they are part of, adapting to changes in substrate availability, the introduction of new reaction steps, or the need for more efficient flux through the pathway. As pathways evolve, enzymes within them may develop new catalytic functions or refine existing ones to better accommodate these changes \citep{noda2018metabolite}. This process frequently involves the co-evolution of enzymes and their substrates. As substrates change—whether due to the introduction of new compounds in the environment or mutations in other metabolic pathways—enzymes may adapt to catalyze reactions with these new substrates, leading to the emergence of entirely new reactions.
Understanding enzyme-substrate interactions, therefore, requires considering their evolutionary dynamics, as these interactions are shaped by the evolutionary history and adaptations of both enzymes and their substrates. This co-evolutionary process is crucial for explaining how enzymes develop new functions and maintain efficiency in response to ongoing changes in their biochemical environment.

To capture the evolutionary dynamics, we introduce the concept of enzyme-reaction co-evolution into EnzymeFlow. We compute the enzyme and reaction evolution by applying multiple sequence alignment (MSA) to enzyme sequences and reaction SMILES, respectively \citep{steinegger2017mmseqs2}. The co-evolution of an enzyme-reaction pair is represented by a matrix \small\( U \in \mathbb{R}^{N_\text{MSA} \times N_\text{token}} \)\normalsize, which combines the MSA results of enzyme sequences and reaction SMILES (illustrated in Fig.~\ref{fig:enzymeflow}(3)(d) \& Fig.~\ref{fig:msa.exp}), where \( N_\text{MSA} \) denotes the number of MSA sequences and \( N_\text{token} \) denotes the length of the MSA alignment preserved. And each element \small\( u^{mn} \in \{1, \dots, 64\} \cup \{\text{\ding{53}}\} \) \normalsize in \( U \) denotes a tokenized character from our co-evolution vocabulary, with additional \text{\ding{53}} indicating the \textit{masking state}.

\textbf{EnzymeFlow on Co-Evolution.}
The flow for co-evolution follows a similar approach to that used for amino acid types and EC-class, treating it as a discrete factor in the design process. The co-evolution is sampled from a Categorical distribution, where each element has probabilities \small\(t  \delta(u^{mn}_1, u^{mn}_t) + (1-t) \delta(\text{\ding{53}}, u^{mn}_t)\)\normalsize. Each element flows independently, reflecting the natural independence of amino acid mutations \citep{boyko2008assessing}. The discrete flow on co-evolution interpolates from the \textit{masking state} \text{\ding{53}} at \(t=0\) to the actual character \(u^{mn}_1\) at \(t=1\). The prediction and loss function are conditioned on the pocket frames and the substrate and product molecules:
\vspace{-0.1cm}
\begin{equation}
\vspace{-0.1cm}
    \mathcal{L}_\text{coevo} = -\sum_{m=1}^{N_\text{MSA}} \sum_{n=1}^{N_\text{token}} \log p_\theta(u^{mn}_1 | {v}_\theta(\mathbf{T}_t, t, l_s, l_p, u^{mn}_t)).
\end{equation}
The model predicts the final co-evolution at \(t=1\) and interpolates to compute its vector field. For co-evolution prediction, we first introduce a co-evolutionary MSA transformer (coEvoFormer) to encode \(U_t\) (implemented in App.~\ref{app:msa.transformer}). The final predicted co-evolution is obtained by computing cross-attention between the encoded enzyme and ligand, and the encoded co-evolution features.

We can therefore express EnzymeFlow with co-evolutionary dynamics for catalytic pocket design as:
\vspace{-0.1cm}
\begin{equation}
\vspace{-0.1cm}
\label{eq:enzyme.gen}
    p_{t}(\mathbf{T}_t, U_t, y_{\text{ec}_t} | \mathbf{T}_1, U_1, y_{\text{ec}_1}, l_s, l_p) = p_t(y_{\text{ec}_t} | y_{\text{ec}_1}, \mathbf{T}_t) \ p_t(U_t | U_1, \mathbf{T}_t) \ p_{t}(\mathbf{T}_t | \mathbf{T}_1, l_s, l_p).
\end{equation}
The final EnzymeFlow model performs flows on protein backbones, amino acid types, EC-class, and enzyme-reaction co-evolution.
Given the SE(3)-invariant prior and the main SE(3)-equivariant network in EnzymeFlow, the pocket generation process is also SE(3)-equivariant (proven in App.~\ref{se3.equivariant}).

\begin{figure*}[t!]
\vspace{-0.5cm}
\centering
{
\includegraphics[width=1\textwidth]{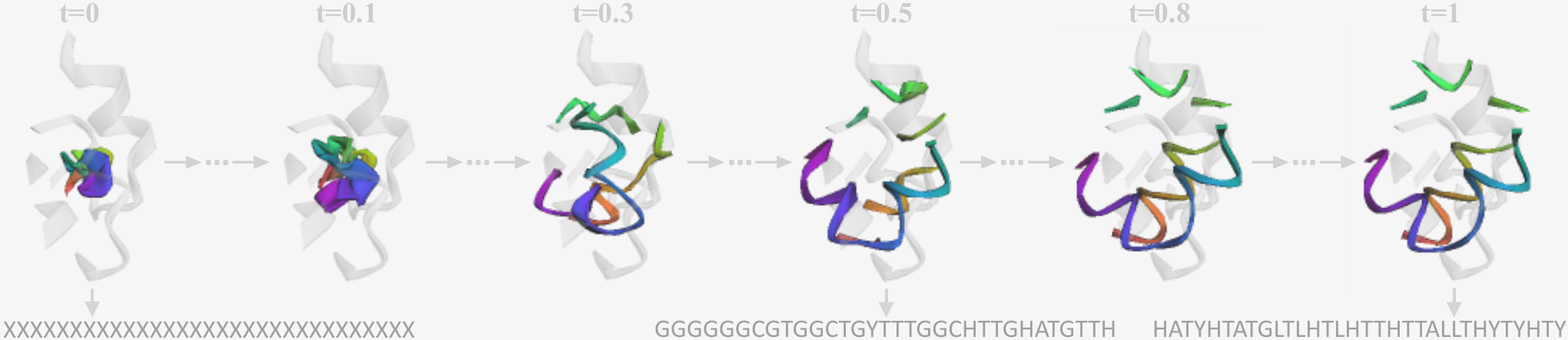}}
{
\vspace{-0.7cm}
  \caption{Catalytic pocket design example using EnzymeFlow (UniProt: Q7U4P2). The pocket generation is conditioned on reaction \tiny\texttt{CN[C@H](C(=O)C)CS.C/C=C$\backslash\backslash$1/C(=C/c2[nH]c(c(c2C)CCC(=O)O)/C=C/2$\backslash\backslash$N=C(C(=C2CCC\\(=O)O)C)C[C@H]2NC(=O)C(=C2C)C=C)/NC(=O)[C@@H]1C $\rightarrow$ CN[C@H](C(=O)C)CSC(C1=C(C)C(=O)N[C@H]1Cc1[nH]c(c(c1C)CCC(=O)\\O)/C=C/1$\backslash\backslash$N=C(C(=C1CCC(=O)O)C)C[C@H]1NC(=O)C(=C1C)C=C)C }\normalsize of EC4 (ligase enzyme), from $t=0$ to $t=1$.}
  \label{fig:generation.example}
  \vspace{-0.5cm}
}
\end{figure*}

\vspace{-0.2cm}
\subsection{Structure-based Hierarchical Pre-training}
\label{sec:pretraining}
\vspace{-0.2cm}
In addition to the standard EnzymeFlow for enzyme pocket design, we propose a hierarchical pre-training strategy to enhance the generalizability of the model across different enzyme categories. The term \textit{hierarchical pre-training} is used because the approach first involves training the flow model to understand protein backbone generation, followed by training it to learn the geometric relationships between proteins and ligand molecules, which form protein binding pockets. After the flow model learns these prior knowledge, we fine-tune it specifically on an enzyme-reaction dataset to generate enzyme catalytic pockets. The term \textit{hierarchical} reflects the progression from protein backbone generation, to protein binding pocket formation, and finally to enzyme catalytic pocket generation.


Specifically, we begin by pre-training the flow model on a protein backbones. Once the model learns it, we proceed to post-train it on a protein-ligands, with the objective of generating binding pockets conditioned on the ligand molecules. Finally, the model is fine-tuned on our EnzymeFlow dataset to generate valid enzyme catalytic pockets for specific substrates and catalytic reactions.

\vspace{-0.2cm}
\subsubsection{Protein Backbone Pre-training}
\vspace{-0.2cm}

The initial step involves pre-training the model on a protein backbone dataset (illustrated in Fig.~\ref{fig:enzymeflow}(1)). We use the backbone dataset discussed in FrameFlow \citep{yim2023fast}. This pre-training focuses solely on SE(3) backbone frames and discrete amino acid types, allowing the flow model to acquire foundational knowledge of protein backbone geometry and structure. 

\vspace{-0.2cm}
\subsubsection{Protein-Ligand Pre-training}
\vspace{-0.2cm}

Following the protein backbone pre-training, we proceed to pre-train the flow model on a protein-ligand dataset (illustrated in Fig.~\ref{fig:enzymeflow}(2)). Specifically, we use PDBBind2020 \citep{wang2004pdbbind}. This pre-training focuses on binding pocket frames, with the flow model conditioned on the 3D representations of ligand molecules \(l\) consisting of \(N_l\) atoms. Additionally, binding affinity \small\(y_\text{kd} \in \mathbb{R}\) \normalsize and atomic-level pocket-ligand distance \small\(D^i \in \mathbb{R}^{4 \times N_l}\) \normalsize for the $i$-th residue frame serve as optimization factors. The parametrization is similar to Eq.~\ref{eq:enzyme.gen}, with conditioning on the ligand molecule as follows:
\vspace{-0.1cm}
\begin{equation}
\vspace{-0.1cm}
    p_{t}(\mathbf{T}_t, y_{\text{kd}} | \mathbf{T}_1, l) = p_t(y_{\text{kd}} | \mathbf{T}_t, l) \ p_{t}(\mathbf{T}_t | \mathbf{T}_1, l).
\end{equation}
In addition to the flow matching losses in Eq.~\ref{eq:backbone.loss}, we introduce a protein-ligand interaction loss to prevent intersection during the binding generation process. Conceptually, this ensures that the generated pocket atoms do not come into contact with the surface of the ligand molecule. Following previous work on protein-ligand binding \citep{lin2022diffbp}, the surface of a ligand 
\small$\{a_j | j \in \mathbb{N}(N_l)\}$\normalsize 
is defined as 
\small$\{a \in \mathbb{R}^3 | S(a) = \gamma \}$\normalsize
, where
\small$S(a)=-\rho \log (\sum_{j=1}^{N_l} \exp (-|a - a_j|^2 / \rho))$\normalsize. 
The interior of the ligand molecule is thus defined by 
\small$\{a \in \mathbb{R}^3 | S(a) < \gamma \}$\normalsize, and the binding pocket atoms are constrained to lie within 
\small$\{a \in \mathbb{R}^3 | S(a) > \gamma \}$\normalsize. 
We also introduce a protein-ligand distance loss to regularize pairwise atomic distances, along with a binding affinity loss to enforce the generation of more valid protein-ligand pairs. These objectives are defined as follows:
\vspace{-0.1cm}
\small
\begin{equation}
\vspace{-0.1cm}
\begin{split}
\mathcal{L}_{\text{inter}} = \sum_{i=1}^{N_r} \max (0,\gamma - S(\hat{A}^i_t)), \
\mathcal{L}_{\text{dist}} & = \sum_{i=1}^{N_r} \frac{\|\mathbf{1}\{D^i_1 < 8\mathring{\text{A}}\} (D^i_1 - \hat{D}^i_t)
\|^2_2}{\sum \mathbf{1}_{D^i_1 < 8\mathring{\text{A}}}}, \ \mathcal{L}_{\text{kd}} = \|y_{\text{kd}} - \hat{y}_{\text{kd}}\|^2,
\end{split}
\end{equation}
\normalsize
where \small$\hat{A}^i \in \mathbb{R}^{ 4 \times 3}$ \normalsize denotes the predicted atomic positions of $i$-th residue frame, \small$\gamma=6$ \normalsize and \small$\rho=2$ \normalsize are hyperparameters, and \small\(\hat{y}_\text{kd}\) \normalsize is the predicted binding affinity for a generated pair. \small\(\hat{D}^i \in \mathbb{R}^{4 \times N_l}\) \normalsize is defined similarly to \small\(D^i\) \normalsize, based on the distance between the predicted atomic positions and ligand positions for the \(i\)-th residue frame. The predicted affinity \small\(\hat{y}_\text{kd}\) \normalsize is obtained by pooling the encoded protein and ligand features.
These additional losses are incorporated to improve the model's generalizability, enforcing more constrained geometries for more valid protein pocket design.

\vspace{-0.2cm}
\section{Experiment — Generating Catalytic Pocket conditioned on Reactions and Substrates}
\vspace{-0.2cm}
EnzymeFlow is essentially a \textit{function-based} protein design model, where the intended function is defined by the reaction the enzyme will catalyze. Here, we demonstrate that EnzymeFlow outperforms current \textit{structure-based} substrate-conditioned protein design models in both the structural and functional aspects, showing its capability and advantage in enzyme catalytic pocket design.

We compare EnzymeFlow with state-of-the-arts representative baselines, including template-matching method DEPACT \citep{chen2022depact}, deep equivariant and iterative refinement model PocketGen \citep{zhang2024pocketgen}, golden-standard diffusion model RFDiffusionAA \citep{krishna2024generalized}, and the most recent PocketFlow\footnote{PocketFlow is not open-sourced yet, we implement and train it on EnzymeFill without fixing the backbones.} \citep{generalized2024zhang}. For RFDiffusionAA-designed pockets, we apply LigandMPNN \citep{dauparas2023atomic} to inverse fold and predict the sequences post-hoc.
We provide the code of EnzymeFlow at \url{https://github.com/WillHua127/EnzymeFlow}.

\begin{table}[t!]
\vspace{-0.5cm}
  \centering
  \caption{EnzymeFlow Evaluation Data Statistics.}
  \resizebox{1.\columnwidth}{!}{%
    \begin{tabular}{c|r|r|r|r|r|r|r|r|r|r|r|r|r}
    \toprule
    \toprule
    \multirow{2}[4]{*}{Data} & \multicolumn{1}{c|}{Pair} & \multicolumn{1}{c|}{Enzyme} & \multicolumn{2}{c|}{Substrate} & \multicolumn{2}{c|}{Product} & \multicolumn{7}{c}{Enzyme Commision} \\
\cmidrule{2-14}          & \multicolumn{1}{c|}{\#pair} & \multicolumn{1}{c|}{\#enzyme} & \multicolumn{1}{c|}{\#substrate} & \multicolumn{1}{c|}{\#avg atom} & \multicolumn{1}{c|}{\#product} & \multicolumn{1}{c|}{\#avg atom} & \multicolumn{1}{c|}{EC1} & \multicolumn{1}{c|}{EC2} & \multicolumn{1}{c|}{EC3} & \multicolumn{1}{c|}{EC4} & \multicolumn{1}{c|}{EC5} & \multicolumn{1}{c|}{EC6} & \multicolumn{1}{c}{EC7} \\
    \midrule
    Raw & 232520 & 97912 & 7259  & 30.81 & 7664  & 30.34 & 44881 (19.30) & 75944 (32.66) & 37728 (16.23) & 47242 (20.32) & 8315 (3.58) & 18281 (7.86) & 129 (0.06) \\
    Train & 53483 & 22350 & 6112  & 30.95 & 6331  & 30.34 & 11674 (21.83) & 18419 (34.44) & 11394 (21.30) & 5555 (10.39) & 2194 (4.10) & 4200 (7.85) & 47 (0.09) \\
    \rowcolor{blue!10}Eval & 100   & 100   & 100   & 30.7  & 94    & 28.84 & 17 (17.00) & 17 (17.00) & 17 (17.00) & 17 (17.00) & 16 (16.00) & 16 (16.00) & 0 (0.00) \\
    \bottomrule
    \bottomrule
    \end{tabular}%
    }
    \vspace{-0.4cm}
  \label{tab:eval.data.stats}%
\end{table}%

\textbf{Evaluation Data.}
We use MMseqs2 to perform clustering with a $10\%$ homology threshold, selecting the center of each cluster as the initial dataset, resulting in a total of $3,417$ pairs. After de-duplicating both repeated substrates and UniProt entries, we are left with $839$ unique enzyme-reaction pairs. We then uniformly sample data across different EC classes, selecting $17$ pairs from EC1 to EC4 classes and $16$ pairs from EC5 and EC6 classes, respectively, resulting in a total of $100$ unique catalytic pockets and $100$ unique reactions. Each enzyme-reaction pair is labeled with a ground-truth EC-class from EC1 to EC6. We present the EC-class distribution in the evaluation set in Tab.~\ref{tab:eval.data.stats}.

\textbf{Reaction-conditioned Generation.}
For pocket design and model sampling, we perform conditional generation on each reaction (or substrate), generating 100 catalytic pockets for each reaction in the evaluation set. We evaluate the generated pockets for their structures and functions (\textit{i.e.,} EC-class).

\textbf{EnzymeFlow Scope.}  
In EnzymeFlow, we adhere to the philosophy that enzyme function dictates its structure. This means that an enzyme folds into a specific 3D shape to fulfill its catalytic role, and the resulting structure can then be inversely folded into a sequence—essentially, \textit{ function $\rightarrow$ structure $\rightarrow$ sequence}. In EnzymeFlow, the enzyme function is defined by the reaction that the enzyme will catalyze (discussed in App.~\ref{app:open.discuss}). We evaluate the structures and functions of the designed pockets.

\vspace{-0.2cm}
\subsection{Catalytic Pocket Structure Evaluation}
\begin{figure*}[t!]
\vspace{-0.6cm}
\centering
{
\includegraphics[width=0.9\textwidth]{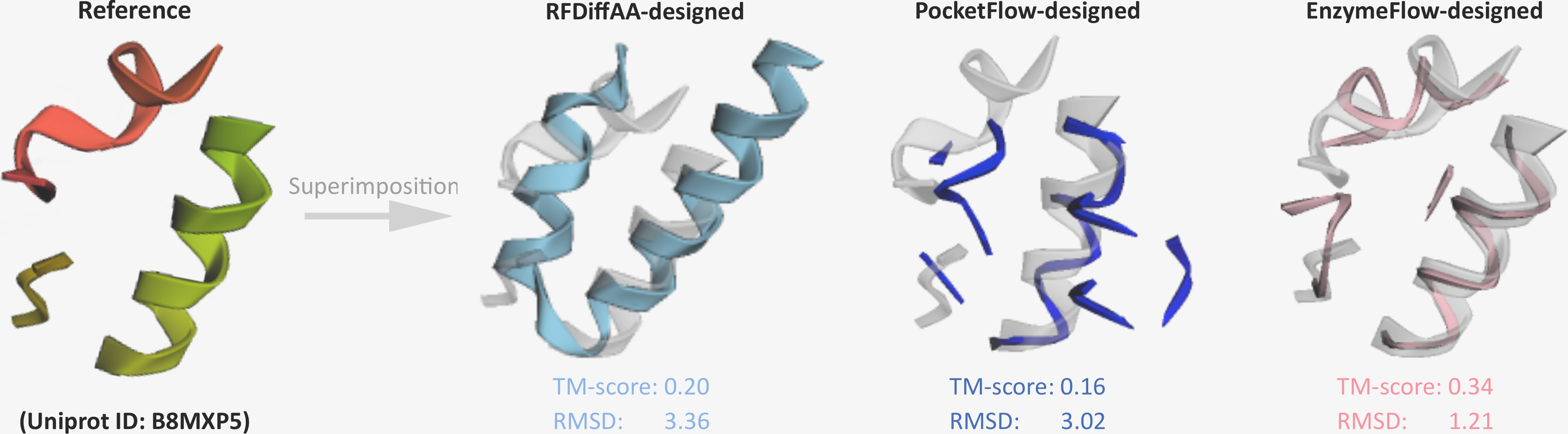}}
{
\vspace{-0.3cm}
  \caption{Case study of catalytic pocket design (UniProt: B8MXP5). We show the reference and designed pockets of different models. The pocket generation is conditioned on reaction \tiny\texttt{OC[C@H]1O[C@@H](Oc2ccccc2/C=C$\backslash\backslash$C(=O)O)[C@@H]([C@H]([C@@H]1O)O)O $\rightarrow$ OC(=O)/C=C$\backslash\backslash$c1ccccc1O }\normalsize of EC3.}
  \label{fig:pocket.example}
  \vspace{-0.5cm}
}
\end{figure*}

\vspace{-0.2cm}
We begin by assessing the structural validity of generated catalytic pockets. While enzyme function determines whether the designed pocket can catalyze a specific reaction, the structure determines whether the substrate conformation can properly bind to the catalytic pocket.
We provide some visual examples of designed pockets in Fig.~\ref{fig:pocket.example} and Fig.~\ref{fig:rfdiff.pocket}.

\begin{table}[ht!]
  \vspace{-0.55cm}
  \centering
  \caption{Evaluation of structural validity of EnzymeFlow- and baseline-generated catalytic pockets. The binding affinities (\texttt{Kd}) and structural confidence (\texttt{chai}) are computed by performing docking on the catalytic pocket and substrate conformation using Vina \citep{trott2010autodock} and Chai \citep{chai2024}, respectively. We highlight top three results in \textbf{bold}, \underline{underline}, and \textit{italic}, respectively.}
  \vspace{0.1cm}
  \resizebox{1.\columnwidth}{!}{%
    \begin{tabular}{c|rrr|rrr|r|r|r|r}
    \toprule
    \toprule
     \multirow{2}[4]{*}{Model} & \multicolumn{3}{c|}{\texttt{cRMSD} ($\downarrow$)} & \multicolumn{3}{c|}{\texttt{TM-score ($\uparrow$)}} & \multicolumn{1}{c|}{\multirow{2}[4]{*}{\texttt{Kd} ($\downarrow$)}} & \multicolumn{1}{c|}{\multirow{2}[4]{*}{\texttt{chai} ($\uparrow$)}} & \multicolumn{1}{c|}{\multirow{2}[4]{*}{\texttt{AAR} ($\uparrow$)}} & \multicolumn{1}{c}{\multirow{2}[4]{*}{\texttt{ECacc} ($\uparrow$)}} \\
    \cmidrule{2-7}       & \multicolumn{1}{c}{Top1} & \multicolumn{1}{c}{Top10} & \multicolumn{1}{c|}{Median} & \multicolumn{1}{c}{Top1} & \multicolumn{1}{c}{Top10} & \multicolumn{1}{c|}{Median} &  &       &       &  \\
    \midrule
    \rowcolor[rgb]{ .655,  .914,  .173} Eval Data & \cellcolor[rgb]{ 1,  1,  1} & \cellcolor[rgb]{ 1,  1,  1}- & \cellcolor[rgb]{ 1,  1,  1} & \cellcolor[rgb]{ 1,  1,  1} & \cellcolor[rgb]{ 1,  1,  1}- & \cellcolor[rgb]{ 1,  1,  1} & \cellcolor[rgb]{ 1,  1,  1}-4.65 & \cellcolor[rgb]{ 1,  1,  1}- & \cellcolor[rgb]{ 1,  1,  1}- & \cellcolor[rgb]{ 1,  1,  1}- \\
    \midrule
    \rowcolor[rgb]{ .855,  .914,  .973} DEPACT & \cellcolor[rgb]{ 1,  1,  1}9.25 & \cellcolor[rgb]{ 1,  1,  1}9.75 & \cellcolor[rgb]{ 1,  1,  1}11.16 & \cellcolor[rgb]{ 1,  1,  1}0.238 & \cellcolor[rgb]{ 1,  1,  1}0.206 & \cellcolor[rgb]{ 1,  1,  1}0.149 & \cellcolor[rgb]{ 1,  1,  1}\underline{-5.46} & \cellcolor[rgb]{ 1,  1,  1}0.125 & \cellcolor[rgb]{ 1,  1,  1}0.112 & \cellcolor[rgb]{ 1,  1,  1}0.149 \\
    \rowcolor[rgb]{ .855,  .914,  .973} PocketGen & \cellcolor[rgb]{ 1,  1,  1}7.65 & \cellcolor[rgb]{ 1,  1,  1}8.14 & \cellcolor[rgb]{ 1,  1,  1}10.45 & \cellcolor[rgb]{ 1,  1,  1}0.260 & \cellcolor[rgb]{ 1,  1,  1}0.233 & \cellcolor[rgb]{ 1,  1,  1}0.193 & \cellcolor[rgb]{ 1,  1,  1}-5.01 & \cellcolor[rgb]{ 1,  1,  1}0.121 & \cellcolor[rgb]{ 1,  1,  1}0.176 & \cellcolor[rgb]{ 1,  1,  1}0.152 \\
    \rowcolor[rgb]{ .855,  .914,  .973} RFDiffusionAA & \cellcolor[rgb]{ 1,  1,  1}9.13 & \cellcolor[rgb]{ 1,  1,  1}9.77 & \cellcolor[rgb]{ 1,  1,  1}11.92 & \cellcolor[rgb]{ 1,  1,  1}0.269 & \cellcolor[rgb]{ 1,  1,  1}0.245 & \cellcolor[rgb]{ 1,  1,  1}0.198 & \cellcolor[rgb]{ 1,  1,  1}\textbf{-12.71} & \cellcolor[rgb]{ 1,  1,  1}\textbf{0.232} & \cellcolor[rgb]{ 1,  1,  1}0.153 & \cellcolor[rgb]{ 1,  1,  1}0.170 \\
    \rowcolor[rgb]{ .855,  .914,  .973} PocketFlow & \cellcolor[rgb]{ 1,  1,  1}7.42 & \cellcolor[rgb]{ 1,  1,  1}8.09 & \cellcolor[rgb]{ 1,  1,  1}10.01 & \cellcolor[rgb]{ 1,  1,  1}0.268 & \cellcolor[rgb]{ 1,  1,  1}\textit{0.260} & \cellcolor[rgb]{ 1,  1,  1}0.197 & \cellcolor[rgb]{ 1,  1,  1}-4.93 & \cellcolor[rgb]{ 1,  1,  1}0.123 & \cellcolor[rgb]{ 1,  1,  1}\textit{0.207} & \cellcolor[rgb]{ 1,  1,  1}0.166 \\
    \midrule
    \rowcolor[rgb]{ 1,  .957,  .941} EnzymeFlow (T=50) & \cellcolor[rgb]{ 1,  1,  1}\textbf{6.94} & \cellcolor[rgb]{ 1,  1,  1}\textbf{7.57} & \cellcolor[rgb]{ 1,  1,  1}\underline{9.04} & \cellcolor[rgb]{ 1,  1,  1}\textbf{0.290} & \cellcolor[rgb]{ 1,  1,  1}\textbf{0.262} & \cellcolor[rgb]{ 1,  1,  1}\textbf{0.209} & \cellcolor[rgb]{ 1,  1,  1}-5.03 & \cellcolor[rgb]{ 1,  1,  1}0.129 & \cellcolor[rgb]{ 1,  1,  1}\textbf{0.216} & \cellcolor[rgb]{ 1,  1,  1}\textbf{0.280} \\
    \rowcolor[rgb]{ 1,  .957,  .941} w/o coevo & \cellcolor[rgb]{ 1,  1,  1}7.02 & \cellcolor[rgb]{ 1,  1,  1}\underline{7.60} & \cellcolor[rgb]{ 1,  1,  1}\textit{9.15} & \cellcolor[rgb]{ 1,  1,  1}\underline{0.288} & \cellcolor[rgb]{ 1,  1,  1}\textit{0.260} & \cellcolor[rgb]{ 1,  1,  1}\textit{0.205} & \cellcolor[rgb]{ 1,  1,  1}-4.86 & \cellcolor[rgb]{ 1,  1,  1}0.123 & \cellcolor[rgb]{ 1,  1,  1}0.196 & \cellcolor[rgb]{ 1,  1,  1}0.246 \\
    \rowcolor[rgb]{ 1,  .957,  .941} w/o pretraining & \cellcolor[rgb]{ 1,  1,  1}\textit{7.01} & \cellcolor[rgb]{ 1,  1,  1}\textit{7.69} & \cellcolor[rgb]{ 1,  1,  1}9.29 & \cellcolor[rgb]{ 1,  1,  1}\textit{0.286} & \cellcolor[rgb]{ 1,  1,  1}\underline{0.261} & \cellcolor[rgb]{ 1,  1,  1}\underline{0.207} & \cellcolor[rgb]{ 1,  1,  1}-4.33 & \cellcolor[rgb]{ 1,  1,  1}\textit{0.134} & \cellcolor[rgb]{ 1,  1,  1}0.202 & \cellcolor[rgb]{ 1,  1,  1}\textit{0.255} \\
    \rowcolor[rgb]{ 1,  .957,  .941} w/o coevo+pretraining & \cellcolor[rgb]{ 1,  1,  1}7.05 & \cellcolor[rgb]{ 1,  1,  1}7.81 & \cellcolor[rgb]{ 1,  1,  1}9.43 & \cellcolor[rgb]{ 1,  1,  1}0.278 & \cellcolor[rgb]{ 1,  1,  1}0.255 & \cellcolor[rgb]{ 1,  1,  1}0.204 & \cellcolor[rgb]{ 1,  1,  1}-4.72 & \cellcolor[rgb]{ 1,  1,  1}0.125 & \cellcolor[rgb]{ 1,  1,  1}0.154 & \cellcolor[rgb]{ 1,  1,  1}0.221 \\
    \rowcolor[rgb]{ 1,  .957,  .941} EnzymeFlow (T=100) & \cellcolor[rgb]{ 1,  1,  1}\underline{6.97} & \cellcolor[rgb]{ 1,  1,  1}\textbf{7.57} & \cellcolor[rgb]{ 1,  1,  1}\textbf{9.02} & \cellcolor[rgb]{ 1,  1,  1}0.283 & \cellcolor[rgb]{ 1,  1,  1}0.258 & \cellcolor[rgb]{ 1,  1,  1}\underline{0.207} & \cellcolor[rgb]{ 1,  1,  1}\textit{-5.31} & \cellcolor[rgb]{ 1,  1,  1}\underline{0.135} & \cellcolor[rgb]{ 1,  1,  1}\underline{0.215} & \cellcolor[rgb]{ 1,  1,  1}\underline{0.273} \\
    \bottomrule
    \bottomrule
    \end{tabular}
    }
    \vspace{-0.2cm}
  \label{tab:structure.result}%
\end{table}%

\textbf{Metrics.}
We use the following metrics to evaluate and compare the structural validity of the generated pocket. 
Constrained-site RMSD (\texttt{cRMSD}): The structural distance between the ground-truth and generated pockets, as proposed in \cite{hayes2024simulating}. \texttt{TM-score}: The topological similarity between the generated and ground-truth pockets in local deviations.
Aggregated Chai Score (\texttt{chai}): The confidence and structural validity of the designed pockets by running Chai \citep{chai2024}. It is calculated as \(0.2 \times \textit{pTM} + 0.8 \times \textit{ipTM} - 100 \times \textit{clash}\), where \textit{pTM} is the predicted template modeling score, \textit{ipTM} is the interface predicted template modeling score (as used in \cite{jumper2021highly}), and the definition of \texttt{chai} is proposed by \cite{chai2024}.
Binding Affinity (\texttt{Kd}): The binding affinity between the generated catalytic pocket and the substrate conformation is computed using AutoDock Vina \citep{trott2010autodock}.
Amino Acid Recovery (\texttt{AAR}): The overlap ratio between the predicted and ground-truth amino acid types in the generated pocket.
Enzyme Commission Accuracy (\texttt{ECacc}): The accuracy of matching the EC-class of generated pockets with the ground-truth EC-class.

\textbf{Results.}
We compare the structural validity between EnzymeFlow- and baseline-generated catalytic pockets in Tab.~\ref{tab:structure.result}. EnzymeFlow and its ablation models outperform baseline models, including leading models like RFDiffusionAA and PocketFlow, with significant improvements in \texttt{cRMSD}, \texttt{TM-score}, and \texttt{ECacc}, and competitive performance in \texttt{AAR}. This demonstrates that EnzymeFlow is capable of generating more structurally valid catalytic pockets, aligning with the enzyme function analysis presented in Fig.~\ref{fig:ec.distribution}. The average improvements over RFDiffusionAA in \texttt{cRMSD}, \texttt{TM-score}, \texttt{AAR}, and \texttt{EC-Acc} are $23.9\%$, $7.8\%$, $41.1\%$, and $64.7\%$, respectively. Additionally, EnzymeFlow slightly outperforms PocketFlow in catalytic-substrate binding, showing improved affinity scores (\texttt{Kd}) and structural confidence (\texttt{chai}) by $2.1\%$ and $9.8\%$, respectively. 

However, EnzymeFlow underperforms RFDiffusionAA in binding scores, reflected by lower affinities and structural confidence. However, considering that the affinities of EnzymeFlow-generated catalytic pockets (-5.03) are close to those of enzyme-reaction pairs in the evaluation set (-4.65), the binding of EnzymeFlow remains acceptable, as enzymes and substrates do not always require tight binding to catalyze reactions because of the kinetic mechanism \citep{cleland1977determining, arcus2020temperature}.

\vspace{-0.2cm}
\subsection{Quantitative Analysis of Enzyme Function}
\vspace{-0.2cm}
The key question is how we can \textit{quantitatively} assess enzyme functions, \textit{i.e.,} catalytic ability, of the generated pockets for a given reaction. To answer this, we perform enzyme function analysis on the designed catalytic pockets. Accurate annotated enzyme function is important for catalytic pocket design because it helps identify the functionality and the active sites that should be preserved or modified to improve catalytic efficiency \citep{rost2002enzyme, barglow2007activity, yu2023enzyme}.

\begin{wrapfigure}{R}{0.48\textwidth}
\vspace{-0.6cm}
  \centering
    \includegraphics[width=0.46\textwidth]{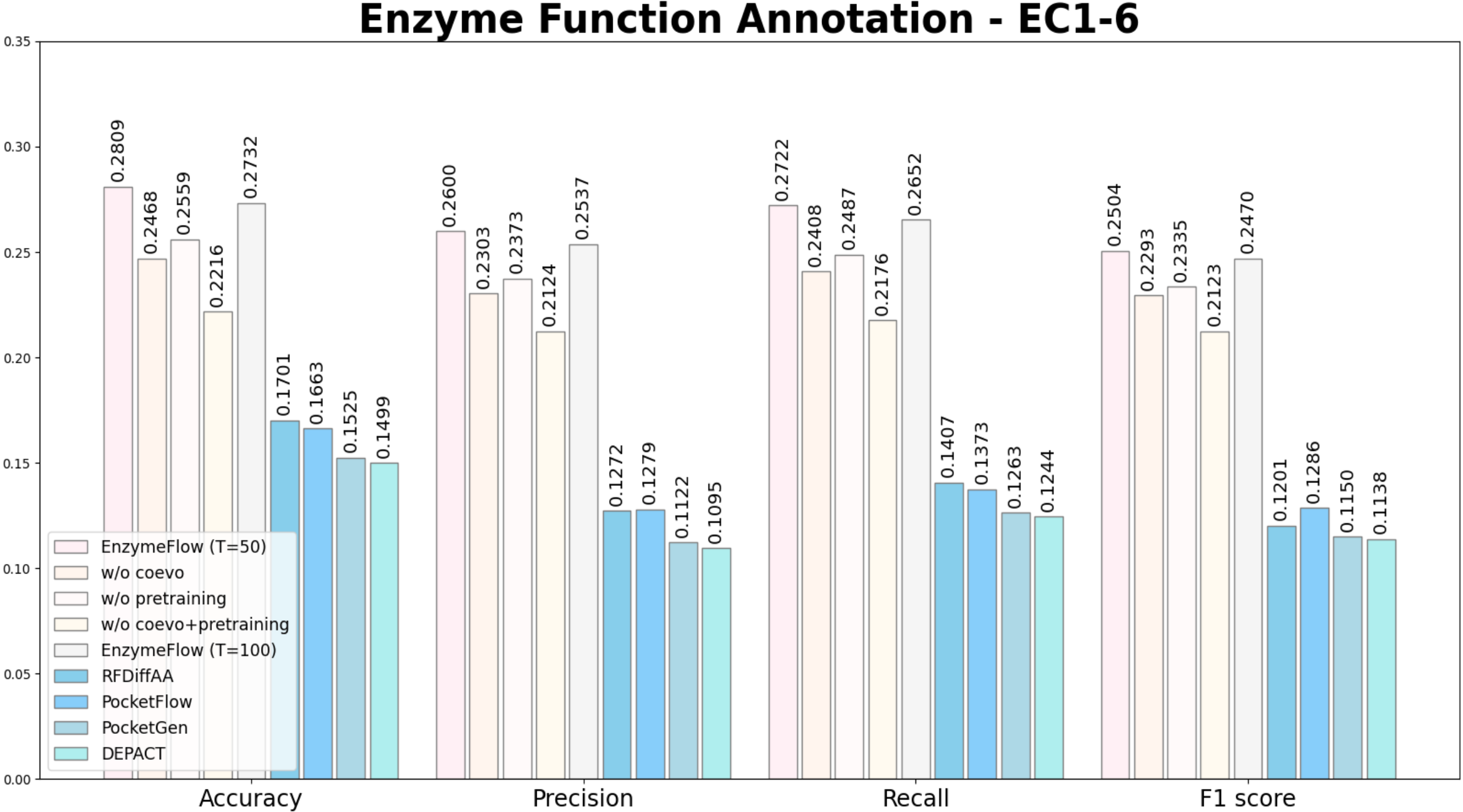}
  \vspace{-0.4cm}
  \caption{Quantitative comparison of annotated enzyme functions between EnzymeFlow- and baseline-generated catalytic pockets across all EC classes, using four multi-label accuracy metrics. \textcolor{Lavender}{Light color represents EnzymeFlow and its ablation models}, \textcolor{SkyBlue}{blue color represents baseline pocket design models}. 
  \vspace{-0.6cm}
  }
   \label{fig:ec.distribution}
\end{wrapfigure}

\textbf{Enzyme Function Comparison.}
In EnzymeFlow, we co-annotate the enzyme function alongside the catalytic pocket design, allowing their functions to directly influence the structure generation. This integration of enzyme function annotation into EnzymeFlow ensures functionality control throughout the design. For baselines that design general proteins rather than enzyme-specific pockets, we perform enzyme function annotation post-hoc using CLEAN \citep{yu2023enzyme} to classify and annotate the EC-class of the generated pockets. After labeling each generated pocket with a EC-class, we compare it to the ground-truth EC-class associated with the actual reaction to compute EC-class accuracy, which quantifies how well the generated pockets align with the intended enzyme functions.

\begin{figure*}[t!]
\vspace{-0.7cm}
\centering
{
\includegraphics[width=1.\textwidth]{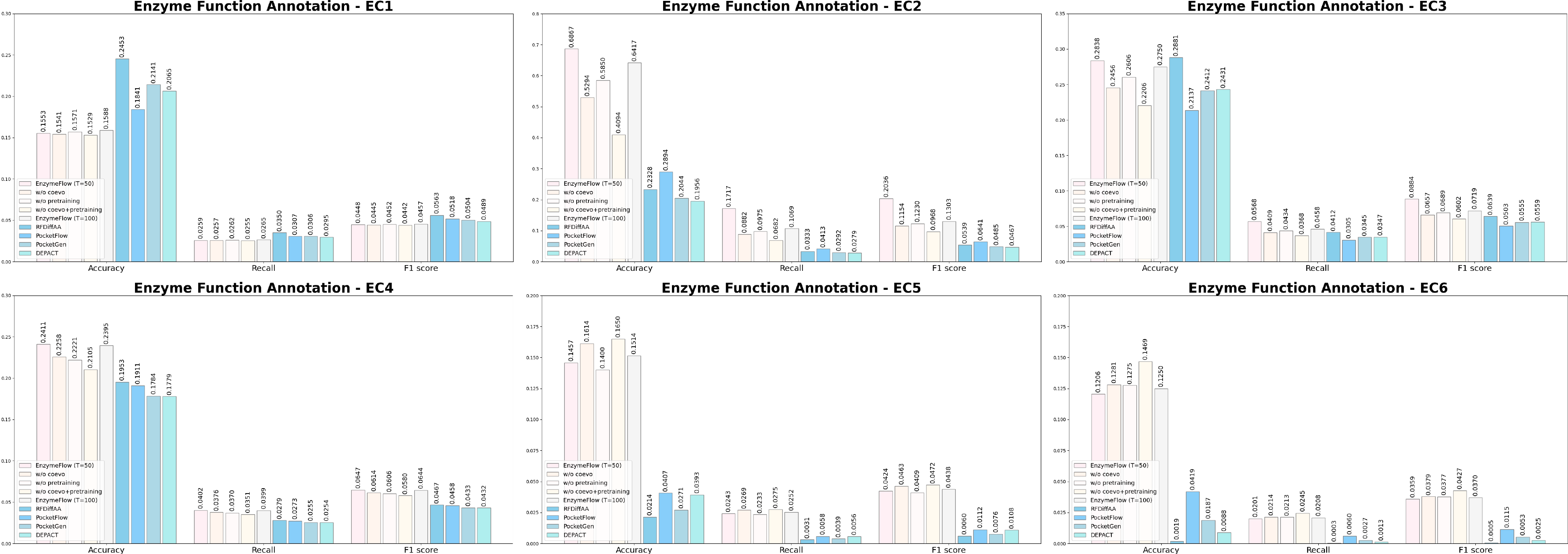}}
{
\vspace{-0.7cm}
  \caption{Quantitative comparison of annotated enzyme functions between EnzymeFlow- and baseline-generated catalytic pockets per EC-class, using accuracy, recall, and F1 score. \textcolor{Lavender}{Light color represents EnzymeFlow and ablation models}, \textcolor{SkyBlue}{blue color represents baseline pocket design models}.}
  \label{fig:ec.per.class}
  \vspace{-0.6cm}
}
\end{figure*}

\textbf{Results.}
We quantitatively compare the annotated enzyme functions between EnzymeFlow- and baseline-generated catalytic pockets across all EC classes in Fig.~\ref{fig:ec.distribution}, and compare the per-class performance in Fig.~\ref{fig:ec.per.class}. These figures allow us to interpret the functions of enzyme catalytic pockets designed by different models. From Fig.~\ref{fig:ec.distribution}, EnzymeFlow and its ablation models achieve the highest values across various multi-label accuracy metrics, including accuracy ($0.2809$), precision ($0.2600$), recall ($0.2722$), and F1 score ($0.2504$), outperforming models like RFDiffusionAA and PocketFlow. 
Additionally, Fig.~\ref{fig:ec.per.class} illustrates per-class enzyme function accuracy, where EnzymeFlow demonstrates strong performance in EC2, EC4, EC5, and EC6, competitive performance in EC3, but slightly weaker performance in EC1 compared to baseline models. Baseline models tend to perform poorly in EC5 and EC6, with per-class occurrence and accuracy showing values close to $0$. In contrast, EnzymeFlow generates more functionally diverse and accurate catalytic pockets, maintaining higher accuracy across different EC classes. 

\begin{wrapfigure}{R}{0.48\textwidth}
\vspace{-0.8cm}
  \centering
    \includegraphics[width=0.46\textwidth]{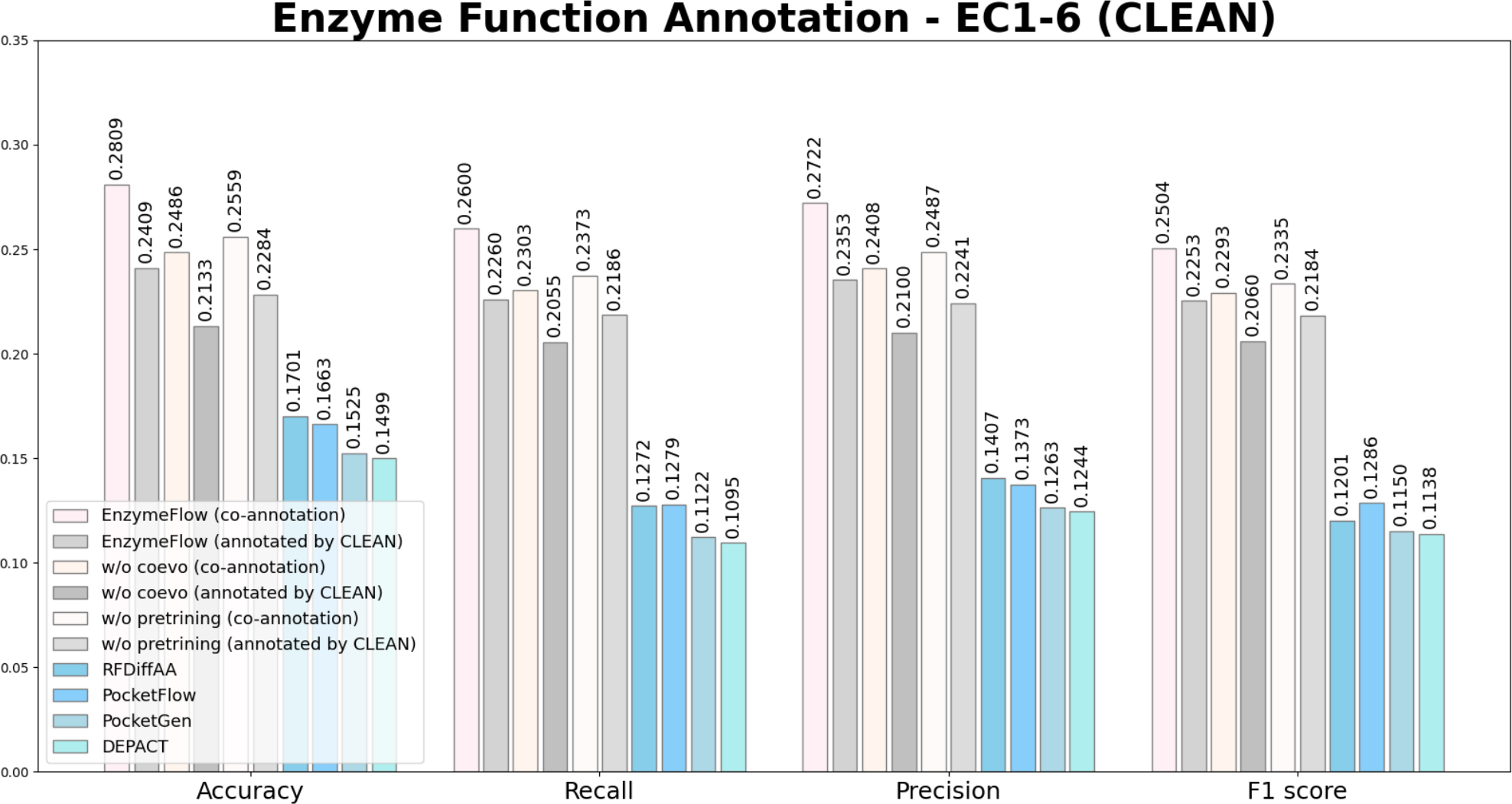}
  \vspace{-0.4cm}
  \caption{Quantitative comparison of annotated enzyme functions between EnzymeFlow- and baseline-generated catalytic pockets across all EC classes, using four multi-label accuracy metrics. \textcolor{Lavender}{Light color represents EnzymeFlow with enzyme function co-annotation}, \textcolor{Gray}{gray color represents EnzymeFlow with enzyme functions annotated by CLEAN post hoc}, \textcolor{SkyBlue}{blue color represents baseline pocket design models with enzyme functions annotated by CLEAN post hoc}.
  \vspace{-0.5cm}
  }
   \label{fig:ec.distribution_clean}
\end{wrapfigure}

Additionally, for a fairer comparison, in Fig.~\ref{fig:ec.distribution_clean}, we compare EnzymeFlow with co-generated enzyme functions, EnzymeFlow with functions annotated post hoc by CLEAN, and baseline models with functions also annotated post hoc by CLEAN. This comparison aims to evaluate the enzyme functions of generated catalytic pockets of different pocket design models using post-hoc function annotation via CLEAN. We observe that EnzymeFlow outperforms the baselines in multi-label accuracy metrics, even when functions are annotated post hoc.

In conclusion, EnzymeFlow generates catalytic pockets that are better compared to other pocket design models, providing more accurate and diverse enzyme functions, which suggests enhanced catalytic potential.
From both functional and structural perspectives, the \textit{function-based, reaction-conditioned} EnzymeFlow outperforms current \textit{structure-based, substrate-conditioned} protein design models in both structural validity and intended function design (catalytic ability). EnzymeFlow leverages enzyme-reaction co-evolution to effectively capture the dynamic changes in catalytic reactions as substrates are transformed into products. This approach enables function-based enzyme design, resulting in the generation of more functionally and structurally valid catalytic pockets for specific reactions.

\vspace{-0.2cm}
\section{Limitation and Future Work}
\vspace{-0.2cm}
\label{sec:future}
EnzymeFlow addresses key challenges in designing enzyme catalytic pockets for specific reactions, but several limitations remain.
The first limitation is that EnzymeFlow currently generates only the catalytic pocket residues, rather than the entire enzyme structure. Ideally, the catalytic pocket should be designed first, followed by the design or reconstruction of the full enzyme structure based on the pocket. While we are developing to use ESM3 \citep{hayes2024simulating} to reconstruct the full enzyme structure based on the designed catalytic pocket (discussed in App.~\ref{app:enzyme.retrieval}), this is not the most ideal solution. ESM3 is not specifically trained for enzyme-related tasks, which may limit its performance in enzyme design. In future versions of EnzymeFlow, we are working to fine-tune large biological models like ESM3 \citep{hayes2024simulating}, RFDiffusionAA \citep{krishna2024generalized}, or Genie2 \citep{lin2024out} to specialize them for enzyme-related tasks, particularly for inpainting functional motifs of enzymes (enzyme catalytic motif scaffolding). Additionally, we aim to create an end-to-end model that combines EnzymeFlow with these large models, enabling catalytic pocket generation and functional motif inpainting in a single step, rather than in a two-step process.
The second limitation, though minor, is that EnzymeFlow currently operates only on enzyme backbones and does not model or generate enzyme side chains. In future work, we plan to incorporate models like DiffPack \citep{zhang2024diffpack} or develop a full-atom model to address this.

\vspace{-0.2cm}
\subsection{Ongoing Work and Broader Impact}
\vspace{-0.2cm}
We introduce EnzymeFlow ongoing work in App.~\ref{app:future.work} and \ref{app:enzyme.retrieval}.
Furthermore, we discuss the scope of EnzymeFlow from structure-based to function-based protein design in greater depth in App.~\ref{app:open.discuss}.

\section*{Acknowledgement}
\vspace{-0.2cm}
This research was supported by the FACS-Acuity Project (No. 10242) and Aureka Bio.

\section*{Reproducibility Statement}
\vspace{-0.2cm}
We provide our code and data examples with demonstrations at \url{https://github.com/WillHua127/EnzymeFlow}. In particular, a Jupyter notebook demonstrating the \textit{de novo} design of enzyme catalytic pockets conditioned on specific reactions is available at \url{https://github.com/WillHua127/EnzymeFlow/blob/main/enzymeflow_demo.ipynb}. For those who prefer not to dive into the full codebase, we have also open-sourced key model components in App.~\ref{app:mol.gnn}, App.~\ref{app:msa.transformer}, and other appendix sections. 

\section*{Collaboration Statement}
\vspace{-0.2cm}
We welcome collaborations, inquiries, questions, and many discussions. Please do not hesitate to reach out to \url{chenqing.hua@mail.mcgill.ca} and \url{shuangjia.zheng@sjtu.edu.cn}. Stay tuned for EnzymeFlow future work!

\newpage

\bibliography{iclr2025_conference}
\bibliographystyle{iclr2025_conference}

\newpage
\appendix

\section{Future Work in Progress: AI-driven Enzyme Design Platform}
\label{app:future.work}
\vspace{-0.2cm}
As discussed in Sec.~\ref{sec:future}, there are several limitations in the current version of EnzymeFlow. Here, we briefly outline the next steps and improvements we are actively working on for the upcoming version. Currently, EnzymeFlow generates only catalytic pocket residues rather than full enzyme structures. Ideally, the catalytic pocket should be designed first, followed by the reconstruction of the full enzyme structure based on the pocket. While we currently use ESM3 \citep{hayes2024simulating} for this reconstruction, this approach is not ideal. Fine-tuning ESM3 or RFDiffusionAA \citep{krishna2024generalized} would be preferable, but unfortunately, training scripts for these wonderful models are not provided, making it impossible to directly fine-tune them on our EnzymeFill dataset.

To address this, we are borrowing concepts from \cite{wang2021deep} and \cite{lin2024out}, which focuses on inpainting proteins and scaffolding functional motifs. We are working to integrate this concept into EnzymeFlow’s pipeline, as part of our primary design. Our goal is to develop an end-to-end automated AI-driven enzyme discovery system that works as follows:
\begin{itemize}
    \item 1. \textbf{Catalytic Pocket Design}: The system will first design enzyme catalytic pockets.
    \item 2. \textbf{Scaffolding Functional Motifs}: Next, it will scaffold the functional motifs to generate full enzyme structures.
    \item 3. \textbf{Substrate Docking}: Using methods like DiffDock \citep{corso2022diffdock}, DynamicBind \citep{lu2024dynamicbind}, or fine-tuned Chai \citep{chai2024} on EnzymeFill, the system will bind substrates to the catalytic pockets.
    \item 4. \textbf{Inverse Folding}: The enzyme-substrate complex will undergo inverse folding using LigandMPNN \citep{dauparas2023atomic}.
    \item 5. \textbf{Computational Screening}: Finally, the system will perform computational screening to select the best-generated enzymes.
\end{itemize}
This entire process is being developed into an integrated, end-to-end solution for AI-driven enzyme design. We are very excited about the potential of this project and look forward to achieving a fully automated enzyme design system in the near future.


\section{Open Discussion: Why is Substrate/Reaction-specified Enzyme Design Needed?}
\label{app:open.discuss}
\vspace{-0.2cm}

EnzymeFlow is unique in its leading approach to function-based \textit{de novo} protein design. Currently, most protein design models, whether focused on backbone generation \citep{yim2023fast, yim2023se, bose2023se, campbell2024generative, krishna2024generalized} or pocket design \citep{zhang2023full, zhang2023efficient, zhang2024pocketgen, generalized2024zhang}, are structure-based. These models aim to design or modify proteins to achieve a specific 3D structure, prioritizing stability, folding, and molecular interactions. The design process typically involves optimizing a protein structure to minimize energy and achieve a stable structural conformation \citep{khoury2014protein, pelay2015structure}.

In contrast, function-based protein design focuses on creating proteins that perform specific biochemical tasks, such as catalysis, signaling, or even binding \citep{martin1998protein, thornton1999protein}. These models are driven by the need for proteins to carry out particular functions rather than adopt a specific 3D structure. Function-based design often targets the active site or binding pockets, optimizing them for specific molecular interactions—in our case, the enzyme’s catalytic pockets.

Our philosophy is that protein function determines its structure, meaning that a protein folds into a specific 3D shape to achieve its intended function, and the resulting structure can then be translated into a proper sequence—essentially, \textit{protein function $\rightarrow$ protein structure $\rightarrow$ protein sequence}. EnzymeFlow follows this philosophy. Specifically, the function of an enzyme is determined by its ability to catalyze a specific reaction or interact with a specific substrate. Therefore, our enzyme pocket design process begins with the reaction or substrate in mind, incorporating reaction/substrate specificity into the generation process. The reaction or substrate represents the functional target for the generated enzyme pockets.

In this approach, EnzymeFlow generates enzyme pocket structures specified for the desired protein function, which contrasts with current generative methods that prioritize structure first. These existing methods operate on the idea that \textit{protein structure $\rightarrow$ protein function $\rightarrow$ protein sequence}. However, proteins should be designed primarily for their functionality, not just their structures. EnzymeFlow’s focus on function-based design could serve as an inspiration for future advancements, leading the way toward more purposeful, function-driven protein design.

\section{Extended Related Work — More Discussion }
\label{app:related.work}
\vspace{-0.2cm}

\subsection{Protein Representation Learning}
\vspace{-0.2cm}
Graph representation learning emerges as a potent strategy for representing and learning about proteins and molecules, focusing on structured, non-Euclidean data \citep{satorras2021n, luan2020complete, luan2022revisiting, hua2022graph, hua2022high, luan2024graph, luan2024heterophilic}. 
In this context, proteins and molecules can be effectively modeled as 2D graphs or 3D point clouds, where nodes correspond to individual atoms or residues, and edges represent interactions between them \citep{gligorijevic2021structure, zhang2022protein, hua2023mudiff, zhang2024deep}. 
Indeed, representing proteins and molecules as graphs or point clouds offers a valuable approach for gaining insights into and learning the fundamental geometric and chemical mechanisms governing protein-ligand interactions.
This representation allows for a more comprehensive exploration of the intricate relationships and structural features within protein-ligand structures \citep{tubiana2022scannet, isert2023structure, zhang2024ecloudgen}.

\subsection{Protein Function Annotation}
\vspace{-0.2cm}
Protein function prediction aims to determine the biological role of a protein based on its sequence, structure, or other features. It is a crucial task in bioinformatics, often leveraging databases such as Gene Ontology (GO), Enzyme Commission (EC) numbers, and KEGG Orthology (KO) annotations \citep{bairoch2000enzyme, gene2004gene, mao2005automated}. Traditional methods like BLAST, PSI-BLAST, and eggNOG infer function by comparing sequence alignments and similarities \citep{altschul1990basic, altschul1997gapped, huerta2019eggnog}. Recently, deep learning has introduced more advanced approaches for protein function prediction \citep{ryu2019deep, kulmanov2020deepgoplus, bonetta2020machine}.
There are two major types of function prediction models, one uses only protein sequence as their input, while the other also uses experimentally-determined or predicted protein structure as input. Typically, these methods predict EC or GO annotations to approximate protein functions, rather than describing the exact catalyzed reaction, which is a limitation of these approaches.

\subsection{Protein Evolution}
\vspace{-0.2cm}
Please refer to Sec.~\ref{sec:2.1}.

\subsection{Generative Models for Protein and Pocket Design}
\vspace{-0.2cm}
Please refer to Sec.~\ref{sec:2.2}.

\section{Co-Evolutionary MSA Transformer}
\label{app:msa.transformer}
\vspace{-0.2cm}
Co-evolution captures the dynamic relationship between an enzyme and its substrate during a catalytic reaction. AlphaFold2 \citep{jumper2021highly} has demonstrated the critical importance of leveraging protein evolution, specifically through multiple sequence alignments (MSA) across protein sequences, to enhance a model's generalizability and expressive power. Previous works, such as MSA Transformer \citep{rao2021msa} and EvoFormer \citep{jumper2021highly}, have focused on encoding and learning protein evolution from MSA results. Proper co-evolution encodings of enzymes and reactions are essential for capturing the dynamic changes that occur during catalytic processes, not only in our EnzymeFlow model but in other models as well.

\begin{figure*}[ht!]
\centering
{
\includegraphics[width=1\textwidth]{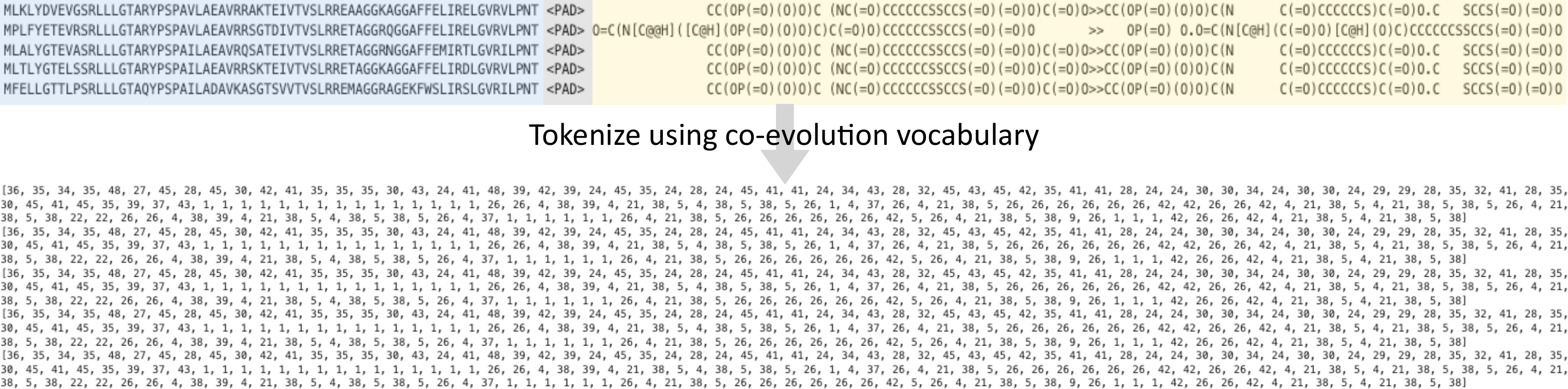}}
\vspace{-0.7cm}
{
  \caption{Enzyme-reaction co-evolution and tokenized representation.}
  \label{fig:msa.exp}
}
\end{figure*}

\subsection{Co-evolution vocabulary}
\label{app:co.evo.dict}
\vspace{-0.2cm}
We provide our co-evolution dictionary for tokenization and encoding following:
\begin{figure*}[ht!]
\centering
{
\includegraphics[width=0.8\textwidth]{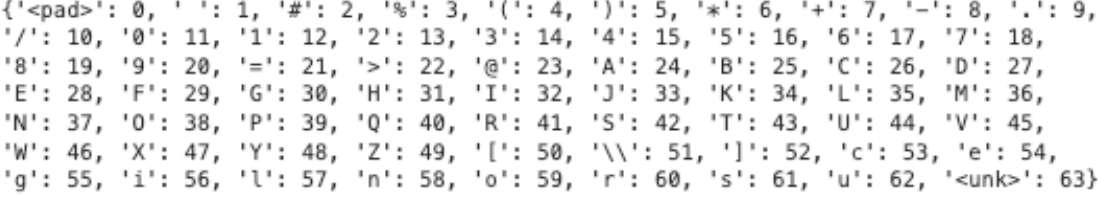}}
\vspace{-0.3cm}
{
  \caption{EnzymeFlow co-evolution dictionary.}
  \label{fig:co.evo.dict}
}
\end{figure*}

\subsection{coEvoFormer Implementation}
\vspace{-0.2cm}
Here, we introduce a new co-evolutionary MSA transformer, coEvoFormer. For details of enzyme-reaction co-evolution, please refer to Sec~\ref{sec:coevo.enzymeflow}.

The code for coEvoFormer follows directly:
\begin{lstlisting}[language=Python, caption=Pytorch Implementation of coEvoFormer.]
import math, copy
import numpy as np

import torch
import torch.nn as nn
import torch.nn.functional as F
from torch.autograd import Variable

## Co-Evolution Transformer (coEvoFormer)

## (12) Layer Norm
class ResidualNorm(nn.Module):
    def __init__(self, size, dropout):
        super(ResidualNorm, self).__init__()
        self.norm = LayerNorm(size)
        self.dropout = nn.Dropout(dropout)

    def forward (self, x, sublayer):
        return x + self.dropout(sublayer(self.norm(x)))


## (11) Residual Norm
class LayerNorm(nn.Module):
    def __init__(self, features, eps=1e-6):
        super(LayerNorm, self).__init__()
        self.a_2 = nn.Parameter(torch.ones(features))
        self.b_2 = nn.Parameter(torch.zeros(features))
        self.eps = eps

    def forward(self, x):
        mean = x.mean(-1, keepdim=True)
        std = x.std(-1, keepdim=True)
        x = self.a_2 * (x - mean) / (std + self.eps) + self.b_2
        return x


## (10) 2-layer MLP
class MLP(nn.Module):
    def __init__(self, model_depth, ff_depth, dropout):
        super(MLP, self).__init__()
        self.w1 = nn.Linear(model_depth, ff_depth)
        self.w2 = nn.Linear(ff_depth, model_depth)
        self.dropout = nn.Dropout(dropout)
        self.silu = nn.SiLU()

    def forward(self, x):
        return self.w2(self.dropout(self.silu(self.w1(x))))


## (9) Attention
def attention(Q,K,V, mask=None):
    dk = Q.size(-1)
    T = (Q @ K.transpose(-2, -1))/math.sqrt(dk)
    if mask is not None:
        T = T.masked_fill_(mask.unsqueeze(1)==0, -1e9)
    T = F.softmax(T, dim=-1)
    return T @ V


## (8) Multi-Head Attention
class MultiHeadAttention(nn.Module):
    def __init__ (self, 
                  num_heads, 
                  embed_dim, 
                  bias=False
                 ):
        super(MultiHeadAttention, self).__init__()
        self.num_heads = num_heads
        self.dk = embed_dim//num_heads
        self.WQ = nn.Linear(embed_dim, embed_dim, bias=bias)
        self.WK = nn.Linear(embed_dim, embed_dim, bias=bias)
        self.WV = nn.Linear(embed_dim, embed_dim, bias=bias)
        self.WO = nn.Linear(embed_dim, embed_dim, bias=bias)

    def forward(self, x, kv, mask=None):
        batch_size = x.size(0)
        Q = self.WQ(x ).view(batch_size, -1, self.num_heads, self.dk).transpose(1,2)
        K = self.WK(kv).view(batch_size, -1, self.num_heads, self.dk).transpose(1,2)
        V = self.WV(kv).view(batch_size, -1, self.num_heads, self.dk).transpose(1,2)

        if mask is not None:
            if len(mask.shape) == 2:
                mask = torch.einsum('bi,bj->bij', mask, mask)
        x = attention(Q, K, V, mask=mask)

        x = x.transpose(1, 2).contiguous().view(batch_size, -1, self.num_heads*self.dk)
        return self.WO(x)


## (7) Positional Embedding
class PositionalEncoding(nn.Module):
    def __init__(self, model_depth, max_len=5000):
        super(PositionalEncoding, self).__init__()

        pe = torch.zeros(max_len, model_depth)
        position = torch.arange(0.0, max_len).unsqueeze(1)
        div_term = torch.exp(torch.arange(0.0, model_depth, 2) *
                             -(math.log(10000.0) / model_depth))
        pe[:, 0::2] = torch.sin(position * div_term)
        pe[:, 1::2] = torch.cos(position * div_term)
        pe = pe.unsqueeze(0)
        self.register_buffer('pe', pe)

    def forward(self, x):
        return x + Variable(self.pe[:, :x.size(1)], requires_grad=False)
        

## (6) Embedding
class Embedding(nn.Module):
    def __init__(self, vocab_size, model_depth):
        super(Embedding, self).__init__()
        self.lut = nn.Embedding(vocab_size, model_depth)
        self.model_depth = model_depth
        self.positional = PositionalEncoding(model_depth)

    def forward(self, x):
        emb = self.lut(x) * math.sqrt(self.model_depth)
        return self.positional(emb)


## (5) Encoder Layer
class EncoderLayer(nn.Module):
    def __init__(self, 
                  n_heads, 
                  model_depth, 
                  ff_depth, 
                  dropout=0.0
                 ):
        super(EncoderLayer, self).__init__()
        self.self_attn = MultiHeadAttention(embed_dim=model_depth, num_heads=n_heads)
        self.resnorm1 = ResidualNorm(model_depth, dropout)
        self.ff = MLP(model_depth, ff_depth, dropout)
        self.resnorm2 = ResidualNorm(model_depth, dropout)

    def forward(self, x, mask):
        x = self.resnorm1(x, lambda arg: self.self_attn(arg, arg, mask))
        x = self.resnorm2(x, self.ff)
        return x

        
## (4) Encoder
class Encoder(nn.Module):
    def __init__ (self, 
                  n_layers, 
                  n_heads, 
                  model_depth, 
                  ff_depth, 
                  dropout
                 ):
        super(Encoder, self).__init__()
        self.layers = nn.ModuleList([EncoderLayer(n_heads, model_depth, ff_depth, dropout) for i in range(n_layers)])
        self.lnorm = LayerNorm(model_depth)

    def forward(self, x, mask):
        for layer in self.layers:
            x = layer(x, mask)
        return self.lnorm(x)


## (3)Generator
class Generator(nn.Module):
    def __init__(self, 
                 model_depth, 
                 vocab_size
                ):
        super(Generator, self).__init__()
        self.ff = nn.Linear(model_depth, vocab_size)

    def forward(self, x):
        return F.log_softmax(self.ff(x), dim=-1)


## (2)coEvoEmbedder
class CoEvoEmbedder(nn.Module):
    def __init__(self, 
                 vocab_size, 
                 n_layers=2, 
                 n_heads=4, 
                 model_depth=64, 
                 ff_depth=64, 
                 dropout=0.0,
                ):
        super(CoEvoFormer, self).__init__()
        
        self.model_depth = model_depth
        self.encoder = Encoder(n_layers=n_layers, 
                               n_heads=n_heads, 
                               model_depth=model_depth, 
                               ff_depth=ff_depth, 
                               dropout=dropout,
                               )
                              
        if vocab_size is not None:
            if isinstance(vocab_size, int):
                self.set_vocab_size(vocab_size)
                
            else:
                self.set_vocab_size(vocab_size[0], vocab_size[1])

    def set_vocab_size(self, src_vocab_size):
        self.src_embedder = Embedding(src_vocab_size, self.model_depth)
        self.generator = Generator(self.model_depth, src_vocab_size)

        for p in self.parameters():
            if p.dim() > 1:
                nn.init.xavier_uniform_(p)

    def forward(self, src, src_mask=None):
        enc_out = self.encoder(self.src_embedder(src), src_mask)
        
        return enc_out


## (1)coEvoFormer
class CoEvoFormer(nn.Module):
    def __init__(self, model_conf):
        super(CoEvoFormer, self).__init__()
        torch.set_default_dtype(torch.float32)
        self._model_conf = model_conf
        self._msa_conf = model_conf.msa

        self.msa_encoder = CoEvoEmbedder(
                            vocab_size=self._msa_conf.num_msa_vocab, 
                            n_layers=self._msa_conf.msa_layers,
                            n_heads=self._msa_conf.msa_heads,
                            model_depth=self._msa_conf.msa_embed_size,
                            ff_depth=self._msa_conf.msa_hidden_size, 
                            dropout=self._model_conf.dropout,
                        )

        self.col_attn = MultiHeadAttention(
                num_heads=self._msa_conf.msa_heads, 
                embed_dim=self._msa_conf.msa_embed_size,
            )

        self.row_attn = MultiHeadAttention(
                num_heads=self._msa_conf.msa_heads, 
                embed_dim=self._msa_conf.msa_embed_size,
            )

    def forward(
        self,
        msa_feature,
        msa_mask=None,
    ):
        bs, n_msa, n_token = msa_feature.size()
        msa_feature = msa_feature.reshape(bs*n_msa, n_token)
        msa_embed = self.msa_encoder(msa_feature).reshape(bs, n_msa, n_token, -1)
        msa_embed = msa_embed.transpose(1, 2).reshape(bs*n_token, n_msa, -1)

        if msa_mask is not None:
            msa_mask = msa_mask.transpose(1, 2).reshape(bs*n_token, n_msa)
            
        msa_embed = self.col_attn(msa_embed, msa_embed, mask=msa_mask).reshape(bs, n_token, n_msa, -1).transpose(1, 2)
        msa_embed = msa_embed.reshape(bs*n_msa, n_token, -1)

        if msa_mask is not None:
            msa_mask = msa_mask.reshape(bs, n_token, n_msa)
            msa_mask = msa_mask.transpose(1, 2).reshape(bs*n_msa, n_token)
            
        msa_embed = self.row_attn(msa_embed, msa_embed, mask=msa_mask).reshape(bs, n_msa, n_token, -1)
        
        return msa_embed
\end{lstlisting}

\section{Molecule GNN}
\label{app:mol.gnn}
\vspace{-0.2cm}

\subsection{3D Molecule GNN}
\vspace{-0.2cm}
The 3D molecule GNN plays a crucial role in EnzymeFlow. During the structure-based hierarchical pre-training, it encodes ligand molecule representations, learning the constrained geometry between protein binding pockets and ligand molecules. This pre-training process makes the 3D molecule GNN transferable. When the flow model is fine-tuned, the 3D molecule GNN is also fine-tuned, transferring its prior knowledge about ligand molecules to substrate molecules in enzyme-catalyzed reactions. This allows for substrate-specific encodings while leveraging the knowledge learned from protein-ligand interactions.

Consider a molecule \( l_s \) with \( N_{l_s} \) atoms; this could be a ligand conformation in a protein-ligand pair or a substrate conformation in an enzyme-substrate pair. The molecule \( l_s \) can be viewed as a set of atomic point clouds in 3D Euclidean space, where each atom is characterized by its atomic type. There is a distance relationship between each atom pair in the point cloud, which can be processed as bonding features. In our 3D molecule GNN, we use a radial basis function to process these pairwise atomic distances, a technique commonly employed to ensure equivariance and invariance in model design \citep{hua2023mudiff, zhang2024deep, zhang2024ecloudgen}. The 3D molecule GNN takes a molecule conformation \( l_s \) as input and outputs an embedded molecule representation \( H_{l_s} \in \mathbb{R}^{N_{l_s} \times D_{H_{l_s}}} \), where \( D_{H_{l_s}} \) denotes the hidden dimension size.

The code for 3D Molecule GNN follows directly:
\begin{lstlisting}[language=Python, caption=Pytorch Implementation of 3D Molecule GNN.]
import math
import numpy as np

import torch
import torch.nn as nn
from torch.nn import functional as F

## (1)3D Molecule GNN
class MolEmbedder3D(nn.Module):
    def __init__(self, model_conf):
        super(MolEmbedder3D, self).__init__()
        torch.set_default_dtype(torch.float32)
        self._model_conf = model_conf
        self._embed_conf = model_conf.embed

        node_embed_dims = self._model_conf.num_atom_type
        node_embed_size = self._model_conf.node_embed_size
        self.node_embedder = nn.Sequential(
            nn.Embedding(node_embed_dims, node_embed_size, padding_idx=0),
            nn.SiLU(),
            nn.Linear(node_embed_size, node_embed_size),
            nn.LayerNorm(node_embed_size),
            )

        self.node_aggregator = nn.Sequential(
            nn.Linear(node_embed_size + self._model_conf.edge_embed_size, node_embed_size),
            nn.SiLU(),
            nn.Linear(node_embed_size, node_embed_size),
            nn.SiLU(),
            nn.Linear(node_embed_size, node_embed_size),
            nn.LayerNorm(node_embed_size),
            )

        self.dist_min = self._model_conf.ligand_rbf_d_min
        self.dist_max = self._model_conf.ligand_rbf_d_max
        self.num_rbf_size = self._model_conf.num_rbf_size
        self.edge_embed_size = self._model_conf.edge_embed_size
        
        self.edge_embedder = nn.Sequential(
            nn.Linear(self.num_rbf_size + node_embed_size + node_embed_size, self.edge_embed_size),
            nn.SiLU(),
            nn.Linear(self._model_conf.edge_embed_size, self._model_conf.edge_embed_size),
            nn.SiLU(),
            nn.Linear(self._model_conf.edge_embed_size, self._model_conf.edge_embed_size),
            nn.LayerNorm(self._model_conf.edge_embed_size),
            )

        mu = torch.linspace(self.dist_min, self.dist_max, self.num_rbf_size)
        self.mu = mu.reshape([1, 1, 1, -1])
        
        self.sigma = (self.dist_max - self.dist_min) / self.num_rbf_size

    # Distance function -- pair-wise distance computation
    def coord2dist(self, coord, edge_mask):
        n_batch, n_atom = coord.size(0), coord.size(1)
        radial = torch.sum((coord.unsqueeze(1) - coord.unsqueeze(2)) ** 2, dim=-1)
        dist = torch.sqrt(
                radial + 1e-10
            ) * edge_mask

        radial = radial * edge_mask
        return radial, dist

    # RBF function -- distance encoding
    def rbf(self, dist):
        dist_expand = torch.unsqueeze(dist, -1)
        _mu = self.mu.to(dist.device)
        rbf = torch.exp(-(((dist_expand - _mu) / self.sigma) ** 2))
        return rbf

    def forward(
        self,
        ligand_atom,
        ligand_pos,
        edge_mask,
    ):
        num_batch, num_atom = ligand_atom.shape

        # Atom Embbedding 
        node_embed = self.node_embedder(ligand_atom)

        # Edge Feature Computation 
        radial, dist = self.coord2dist(
                            coord=ligand_pos, 
                            edge_mask=edge_mask,
                        )
        edge_embed = self.rbf(dist) * edge_mask[..., None]
        src_node_embed = node_embed.unsqueeze(1).repeat(1, num_atom, 1, 1)
        tar_node_embed = node_embed.unsqueeze(2).repeat(1, 1, num_atom, 1)
        edge_embed = torch.cat([src_node_embed, tar_node_embed, edge_embed], dim=-1)
        
        # Edge Embedding
        edge_embed = self.edge_embedder(edge_embed.to(torch.float))

        # Message-Passing
        src_node_agg = (edge_embed.sum(dim=1) / (edge_mask[..., None].sum(dim=1)+1e-10)) * ligand_atom.clamp(max=1.)[..., None]
        src_node_agg = torch.cat([node_embed, src_node_agg], dim=-1)
        
        # Residue Connection
        node_embed = node_embed + self.node_aggregator(src_node_agg)

        return node_embed, edge_embed
\end{lstlisting}

\subsection{2D Molecule GNN}
\vspace{-0.2cm}
Like the 3D molecule GNN, the 2D molecule GNN is also important in our EnzymeFlow implementation. In an enzyme-catalyzed reaction, the substrate molecule is transformed into a product molecule, with enzyme-substrate interactions driving this chemical transformation. The 2D molecule GNN plays a key role in modeling and encoding this transformation during the catalytic process, making it equally important as our use of co-evolutionary dynamics. While the 3D molecule GNN encodes the substrate, the 2D molecule GNN encodes the product, guiding the design of the enzyme catalytic pocket.

Consider a product molecule \( l_p \) with \( N_{l_p} \) atoms in a catalytic reaction. This molecule can be represented as a graph, where nodes correspond to atoms and edges represent bonds. In our 2D molecule GNN, we use fingerprints with attention mechanisms \citep{xiong2019pushing} to facilitate message passing between atoms, enabling effective communication across the molecule. The 2D molecule GNN takes this molecular graph \( l_p \) as input and outputs an embedded molecule representation \( H_{l_p} \in \mathbb{R}^{N_{l_p} \times D_{H_{l_p}}} \), where \( D_{H_{l_p}} \) denotes the hidden dimension size.

The code for 2D Molecule GNN follows directly:
\begin{lstlisting}[language=Python, caption=Pytorch Implementation of 2D Molecule GNN.]
import torch
import torch.nn as nn
from torch_geometric.nn.models import AttentiveFP

## (1)2D Molecule GNN
class MolEmbedder2D(nn.Module):
    def __init__(self, model_conf):
        super(MolEmbedder2D, self).__init__()
        torch.set_default_dtype(torch.float32)
        self._model_conf = model_conf

        self.node_embed_dims = self._model_conf.mpnn.mpnn_node_embed_size
        self.edge_embed_dims = self._model_conf.mpnn.mpnn_edge_embed_size

        self.node_embedder = nn.Sequential(
            nn.Embedding(self._model_conf.num_atom_type, self.node_embed_dims),
            nn.SiLU(),
            nn.Linear(self.node_embed_dims, self.node_embed_dims),
            nn.LayerNorm(self.node_embed_dims),
            )

        self.edge_embedder = nn.Sequential(
            nn.Embedding(self._model_conf.mpnn.num_edge_type, self.edge_embed_dims),
            nn.SiLU(),
            nn.Linear(self.edge_embed_dims, self.edge_embed_dims),
            nn.LayerNorm(self.edge_embed_dims),
            )

        # Message Passing with Atttention and Fingerprint
        self.mpnn = AttentiveFP(
                    in_channels=self.node_embed_dims,
                    hidden_channels=self.node_embed_dims,
                    out_channels=self.node_embed_dims,
                    edge_dim=self.edge_embed_dims,
                    num_layers=self._model_conf.mpnn.mpnn_layers,
                    num_timesteps=self._model_conf.mpnn.n_timesteps,
                    dropout=self._model_conf.mpnn.dropout,
                )

    # Dense Edge Matrix to Sparse Edge Matrix
    def dense_to_sparse(
        self,
        mol_atom,
        mol_edge,
        mol_edge_feat,
        mol_atom_mask,
        mol_edge_mask,
    ):
        mol_atom_list = mol_atom[mol_atom_mask]
        mol_edge_feat_list = mol_edge_feat[mol_edge_mask]

        if mol_edge.size(dim=1) == 2:
            mol_edge = mol_edge.transpose(1,2)
        mol_edge_list = [edge[mask] for edge, mask in zip(mol_edge, mol_edge_mask)]
        
        n_nodes = mol_atom_mask.sum(dim=1, keepdim=True)
        cum_n_nodes = torch.cumsum(n_nodes, dim=0)
        new_mol_edge_list = [mol_edge_list[0]]
        for edge, size in zip(mol_edge_list[1:], cum_n_nodes[:-1]):
            new_mol_edge = edge + size
            new_mol_edge_list.append(new_mol_edge)
            
        new_mol_edge_list = torch.cat(new_mol_edge_list, dim=0)

        if new_mol_edge_list.size(dim=1) == 2:
            new_mol_edge_list = new_mol_edge_list.transpose(1,0)

        idx = 0
        batch_mask = []
        for size in n_nodes:
            batch_mask.append(torch.zeros(size, dtype=torch.long) + idx)
            idx += 1
        batch_mask = torch.cat(batch_mask).to(mol_atom.device)
        
        return mol_atom_list, new_mol_edge_list, mol_edge_feat_list, batch_mask

    def forward(
        self,
        mol_atom,
        mol_edge,
        mol_edge_feat,
        mol_atom_mask,
        mol_edge_mask,
    ):
        n_batch = mol_atom.size(0)
        
        mol_atom_mask = mol_atom_mask.bool()
        mol_edge_mask = mol_edge_mask.bool()
        mol_atom, mol_edge, mol_edge_feat, batch_mask = self.dense_to_sparse(mol_atom, mol_edge, mol_edge_feat, mol_atom_mask, mol_edge_mask)
        assert mol_edge.size(1) == mol_edge_feat.size(0)

        # Atom Embedding
        mol_atom = self.node_embedder(mol_atom)

        # Edge Embedding
        mol_edge_feat = self.edge_embedder(mol_edge_feat)

        # Message-Passing
        mol_rep = self.mpnn(mol_atom, mol_edge, mol_edge_feat, batch_mask)
        
        return mol_rep
\end{lstlisting}

\section{Vector Field Computation and Sampling}
\label{app:vec.field}
\vspace{-0.2cm}
Here, we describe how to compute vectors fields and perform sampling for catalytic pocket residues frames, EC-class, as well as the enzyme-reaction co-evolution.

\subsection{Background}
\vspace{-0.2cm}
\textbf{Catalytic Pocket Frame.}
Please refer to Sec.~\ref{sec:general.enzymeflow}.

\textbf{EC-Class.} 
Please refer to Sec.~\ref{sec:general.enzymeflow}.

\textbf{Co-evolution.} 
Please refer to Sec.~\ref{sec:coevo.enzymeflow}.

\textbf{Vector Field.}
Flow matching describes a process where a flow transforms a simple distribution \( p_0 \) into the target data distribution \( p_1 \) \citep{lipman2022flow}. The goal in flow matching is to train a neural network \( v_\theta(\epsilon_t, t) \) that approximates the vector field \( u_t(\epsilon) \), which measures the transformation of the distribution \( p_t(\epsilon_t) \) as it evolves toward \( p_1(\epsilon_t) \) over time \( t \in [0, 1) \). The process is optimized using a regression loss defined as \( \mathcal{L}_{\text{FM}} = \mathbb{E}_{t\sim \mathcal{U}[0,1], p_t(\epsilon_t)} \|v_\theta(\epsilon_t, t) - u_t(\epsilon)\|^2 \).
However, directly computing \( u_t(\epsilon) \) is often intractable in practice. Instead, a conditional vector field \( u_t(\epsilon | \epsilon_1) \) is defined, and the conditional flow matching objective is computed as \( \mathcal{L}_{\text{CFM}} = \mathbb{E}_{t\sim \mathcal{U}[0,1], p_t(\epsilon_t)} \|v_\theta(\epsilon_t, t) - u_t(\epsilon | \epsilon_1)\|^2 \). Notably, \( \nabla_\theta \mathcal{L}_\text{FM} = \nabla_\theta \mathcal{L}_\text{CFM} \).

During inference or sampling, an ODEsolver, \textit{e.g.}, Euler method, is typically used to solve the ODE governing the flow, expressed as \( \epsilon_1 = \texttt{ODEsolver}(\epsilon_0, v_\theta, 0, 1) \), where \( \epsilon_0 \) is the initial data and \( \epsilon_1 \) is the generated data.
In actual training, rather than directly predicting the vector fields, it is more common to use the neural network to predict the final state at \( t=1 \), then interpolates to calculate the vector fields. This approach has been shown to be more efficient and effective for network optimization \citep{yim2023fast, bose2023se, campbell2024generative}.

\subsection{Continuous Variable Trajectory}
\vspace{-0.3cm}
Given the predictions for translation \(\hat{x}_1\) and rotation \(\hat{r}_1\) at \(t=1\), we interpolate and their corresponding vector fields are computed as follows:
\begin{equation}
    {v}_\theta(x_t, t) = \frac{\hat{x}_1 - x_t}{1-t}, \quad {v}_\theta(r_t, t) = \frac{\log_{r_t} \hat{r}_1}{1-t}.
\end{equation}
The sampling or trajectory can then be computed using Euler steps with a step size \(\Delta t\), as follows:
\begin{equation}
    x_{t + \Delta t} = x_t + {v}_\theta(x_t, t) \cdot \Delta t, \quad r_{t + \Delta t} = r_t + {v}_\theta(r_t, t) \cdot \Delta t,
\end{equation}
where the prior of $x_0, r_0$ are chosen as the uniform distribution on $\mathbb{R}^3$ and SO(3), respectively.

\subsection{Discrete Variable Trajectory}
\vspace{-0.3cm}
For the discrete variables, including amino acid types, EC-class, and co-evolution, we follow \cite{campbell2024generative} to use continuous time Markov chains (CTMC). 

\textbf{Continuous Time Markov Chain.}
A sequence trajectory $\epsilon_t$ over time $t \in [0, 1]$ that follows a CTMC alternates between resting in its current state and
periodically jumping to another randomly chosen state. The frequency and destination of the jumps are determined by the rate matrix $R_t \in \mathbb{R}^{N \times N}$ with the constraint its off-diagonal elements are non-negative. The probability of $\epsilon_t$ jumping to a different state $s$ follows $R_t(\epsilon_t, s)\mathrm{d}t$ for the next infinitesimal time step $\mathrm{d}t$. We can express the transition probability as
\begin{equation}
    p_{t + \mathrm{d}t}(s | \epsilon_t) = \delta \{\epsilon_t, s\} + R_t(\epsilon_t, s)\mathrm{d}t,
\end{equation}
where \(\delta(\text{a}, \text{b})\) is the Kronecker delta, equal to \(1\) if \(\text{a} = \text{b}\) and \(0\) if \(\text{a} \neq \text{b}\), and $R_t(\epsilon_t, \epsilon_t) = -\sum_{\gamma \neq \epsilon} (\epsilon_t, \gamma)$ \citep{campbell2024generative}. Therefore, $p_{t+\mathrm{d}t}$ is a Categorical distribution with probabilities \(\delta(\epsilon_t, \cdot) + R_t(\epsilon_t, \cdot)\mathrm{d}t\) with notation $s \sim \text{Cat}(\delta(\epsilon_t, s) + R_t(\epsilon_t, s)\mathrm{d}t)$.

For finite time intervals $\Delta t$, a sequence trajectory can be
simulated with Euler steps following:
\begin{equation}
    \epsilon_{t+\Delta t} \sim \text{Cat}(\delta(\epsilon_t, \epsilon_{t+\Delta t}) + R_t(\epsilon_t, \epsilon_{t+\Delta t})\Delta t).
\end{equation}
The rate matrix $R_t$ along with an initial distribution $p_0$ define CTMC.
Furthermore, the probability flow $p_t$ is the marginal distribution of $\epsilon_t$ at every time $t$, and we say the rate matrix $R_t$ generates $p_t$ if $\partial_t p_t = R^T_tp_t, \forall t \in [0,1]$.

In the actual training, \cite{campbell2024generative} show that we can train a neural network to approximate the true denoising distribution using the standard cross-entropy:
\begin{equation}
    \mathcal{L}_\text{CE} = \mathbb{E}_{t\sim \mathcal{U}[0,1], p_t(\epsilon_t)} [\log p_\theta(\epsilon_1 | \epsilon_t)],
\end{equation}
which leads to our neural network objectives for amino acid types, EC-class, and co-evolution as:
\small
\begin{equation}
\begin{split}
    &\mathcal{L}_\text{aa} = \mathbb{E}_{t\sim \mathcal{U}[0,1], p_t(c_t)} [\log p_\theta(c_1 | c_t)], \mathcal{L}_\text{ec} = \mathbb{E}_{t\sim \mathcal{U}[0,1], p_t(y_{\text{ec}_t})} [\log p_\theta(y_{\text{ec}_1} | y_{\text{ec}_t})], \\ &\mathcal{L}_\text{coevo} = \mathbb{E}_{t\sim \mathcal{U}[0,1], p_t(u_t)} [\log p_\theta(u_1 | u_t)].
\end{split}
\end{equation}
\normalsize

\textbf{Rate Matrix for Inference.}
The conditional rate matrix $R_t(\epsilon_t, s|s_1)$ generates the conditional flow $p_t(\epsilon_t | \epsilon_1)$. 
And $R_t(\epsilon_t, s) = \mathbb{E}_{p_1(\epsilon_1 | \epsilon_t)}[R_t(\epsilon_t, s | \epsilon_1)]$, for which the expectation is taken over $p_1(\epsilon_1|\epsilon_t) = \frac{p_t(\epsilon_t|\epsilon_1)p_1(\epsilon_1)}{p_t(\epsilon_t)}$. 
With the conditional rate matrix, the sampling can be performed:
\begin{equation}
\begin{split}
    & R_t(\epsilon_t, \cdot) \leftarrow \mathbb{E}_{p_1(\epsilon_1|\epsilon_t)}[R_t(\epsilon_t, \cdot | \epsilon_1)], \\
    & \epsilon_{t+\Delta t} \sim \text{Cat}(\delta(\epsilon_t, \epsilon_{t+\Delta t}) + R_t(\epsilon_t, \epsilon_{t+\Delta t})\Delta t).
\end{split}
\end{equation}
The rate matrix generates the probability flow for discrete variables.

\cite{campbell2024generative} define the conditional rate matrix starting with
\begin{equation}
    R_t(\epsilon_t, s | \epsilon_t) = \frac{\text{ReLU}(\partial_t p_t(s | \epsilon_1)- \partial_t p_t(\epsilon_t | \epsilon_1))}{N \cdot p_t(\epsilon_t | \epsilon_1)}.
\end{equation}
In practice, the closed-form of conditional rate matrix with \textit{masking state} \ding{53} is defined as:
\begin{equation}
    R_t(\epsilon_t, s| \epsilon_1) = \frac{\delta(\epsilon_1, s)}{1-t} \delta(\epsilon_t, \text{\ding{53}}).
\end{equation}

With the definition of the conditional rate matrix $R_t(\epsilon_t, s| \epsilon_1)$, we can perform sampling and inference for amino acid types, EC-class, and co-evolution following:
\begin{equation}
\begin{split}
    &c_{t + \Delta t} \sim \text{Cat}(\delta(c_t, c_{t+\Delta t}) + R_t(c_t, c_{t+\Delta t}| v_\theta(c_t, t)) \cdot \Delta t), \\
    &y_{\text{ec}_{t + \Delta t}} \sim \text{Cat}(\delta(y_{\text{ec}_{t}}, y_{\text{ec}_{{t+\Delta t}}}) + R_t(y_{\text{ec}_{t}}, y_{\text{ec}_{{t+\Delta t}}}| v_\theta(y_{\text{ec}_{t}}, t)) \cdot \Delta t), \\
    &u_{t + \Delta t} \sim \text{Cat}(\delta(u_t, u_{t+\Delta t}) + R_t(u_t, u_{t+\Delta t}| v_\theta(u_t, t)) \cdot \Delta t).
\end{split}
\end{equation}

\section{EnzymeFlow SE(3)-equivariance}
\label{se3.equivariant}
\vspace{-0.2cm}
\textbf{Theorem.}  
\textit{Let \(\phi\) denote an SE(3) transformation. The catalytic pocket design in EnzymeFlow, represented as \( p_\theta(\mathbf{T} | l_s) \), is SE(3)-equivariant, meaning that \( p_\theta(\phi(\mathbf{T}) | \phi(l_s)) = p_\theta(\mathbf{T} | l_s) \), where \(\mathbf{T}\) represents the generated catalytic pocket, and \(l_s\) denotes the substrate conformation.}

\textit{Proof.}  
Given an SE(3)-invariant prior, such that \( p(\mathbf{T}_0, l_s) = p(\phi(\mathbf{T}_0), \phi(l_s)) \), and an SE(3)-equivariant transition state for each time step \( t \) via an SE(3)-equivariant neural network, such that \( p_\theta(\mathbf{T}_{t+\Delta t}, l_s) = p_\theta(\phi(\mathbf{T}_{t+\Delta t}), \phi(l_s)) \), it follows that for the total time steps \( T \), we have:
\begin{equation}
\begin{split}
p_\theta(\phi(\mathbf{T}_1) | \phi(l_s)) &= \int p_\theta(\phi(\mathbf{T}_0, l_s)) \prod_{n=0}^{T-1} p_\theta (\phi(\mathbf{T}_{n\Delta t + \Delta t}, l_s)|\phi(\mathbf{T}_{n\Delta t}, l_s))\\ 
&= \int p_\theta(\mathbf{T}_0, l_s) \prod_{n=0}^{T-1} p_\theta (\phi(\mathbf{T}_{n\Delta t + \Delta t}, l_s)|\phi(\mathbf{T}_{n\Delta t}, l_s))\\
&= \int p_\theta(\mathbf{T}_0, l_s) \prod_{n=0}^{T-1} p_\theta (\mathbf{T}_{n\Delta t + \Delta t}, l_s|\mathbf{T}_{n\Delta t}, l_s)\\
&= p_\theta(\mathbf{T}_1 | l_s).  \ \ \ \ \ \ \ \ \ \ \ \ \ \ \ \ \ \ \ \ \ \ \ \ \ \ \ \ \ \ \ \ \ \ \ \ \ \ \ \ \ \ \ \ \ \ \ \ \ \ \ \ \ \ \ \ \ \ \ \ \ \ \ \  \square
\end{split}
\end{equation}

\section{EnzymeFill Dataset Statistics}
\label{app:data.stats}
\vspace{-0.2cm}
\textbf{Data Source.}
Please refer to Sec.~\ref{sec:enzymeflow.dataset}.

\begin{table}[ht!]
  \centering
  \resizebox{1.\columnwidth}{!}{%
    \begin{tabular}{c|r|r|r|r|r|r|r|r|r|r|r|r|r}
    \toprule
    \multirow{2}[4]{*}{Data} & \multicolumn{1}{c|}{Reaction} & \multicolumn{1}{c|}{Enzyme} & \multicolumn{2}{c|}{Substrate} & \multicolumn{2}{c|}{Product} & \multicolumn{7}{c}{Enzyme Commission Class} \\
\cmidrule{2-14}          & \multicolumn{1}{c|}{\#reaction} & \multicolumn{1}{c|}{\#enzyme} & \multicolumn{1}{c|}{\#substrate} & \multicolumn{1}{c|}{\#avg atom} & \multicolumn{1}{c|}{\#product} & \multicolumn{1}{c|}{\#avg atom} & \multicolumn{1}{c|}{EC1} & \multicolumn{1}{c|}{EC2} & \multicolumn{1}{c|}{EC3} & \multicolumn{1}{c|}{EC4} & \multicolumn{1}{c|}{EC5} & \multicolumn{1}{c|}{EC6} & \multicolumn{1}{c}{EC7} \\
    \midrule
    Rawdata & 232520 & 97912 & 7259  & 30.81 & 7664  & 30.34 & 44881 (19.30$\%$) & 75944 (32.66$\%$) & 37728 (16.23$\%$) & 47242 (20.32$\%$) & 8315 (3.58$\%$) & 18281 (7.86$\%$) & 129 (0.06$\%$) \\
    \midrule
    40$\%$ Homo &  19379 & 6922  & 4798  & 31.06 & 4897  & 30.24 & 4754 (24.53$\%$) & 5857 (30.22$\%$) & 4839 (24.97$\%$) & 1764 (9.10$\%$) & 759 (3.92$\%$) & 1379 (7.12$\%$) & 27 (0.14$\%$) \\
    50$\%$ Homo & 34750 & 13442 & 5675  & 31.45 & 5871  & 30.75 & 8184 (23.55$\%$) & 11174 (32.16$\%$) & 8050 (23.17$\%$) & 3203 (9.22$\%$) & 1357 (3.91$\%$) & 2752 (7.92$\%$) & 30 (0.09$\%$) \\
    60$\%$ Homo & 53483 & 22350 & 6112  & 30.95 & 6331  & 30.34 & 11674 (21.83$\%$) & 18419 (34.44$\%$) & 11394 (21.30$\%$) & 5555 (10.39$\%$) & 2194 (4.10$\%$) & 4200 (7.85$\%$) & 47 (0.09$\%$) \\
    80$\%$ Homo & 100925 & 43458 & 6619  & 30.46 & 6943  & 29.95 & 21308 (21.11$\%$) & 34344 (34.03$\%$) & 18925 (18.75$\%$) & 14010 (13.88$\%$) & 3901 (3.87$\%$) & 8371 (8.29$\%$) & 66 (0.07$\%$) \\
    90$\%$ Homo &  132047 & 55697 & 6928  & 30.32 & 7298  & 29.81 & 28833 (21.84$\%$) & 43287 (32.78$\%$) & 23989 (18.17$\%$) & 20070 (15.20$\%$) & 5015 (3.80$\%$) & 10766 (8.15$\%$) & 87 (0.07$\%$) \\
    \bottomrule
    \end{tabular}%
    }
    \caption{EnzymeFill Dataset Statistics.}
  \label{tab:data.stats}%
\end{table}%

\begin{figure*}[ht!]
\centering
{
\includegraphics[width=.9\textwidth]{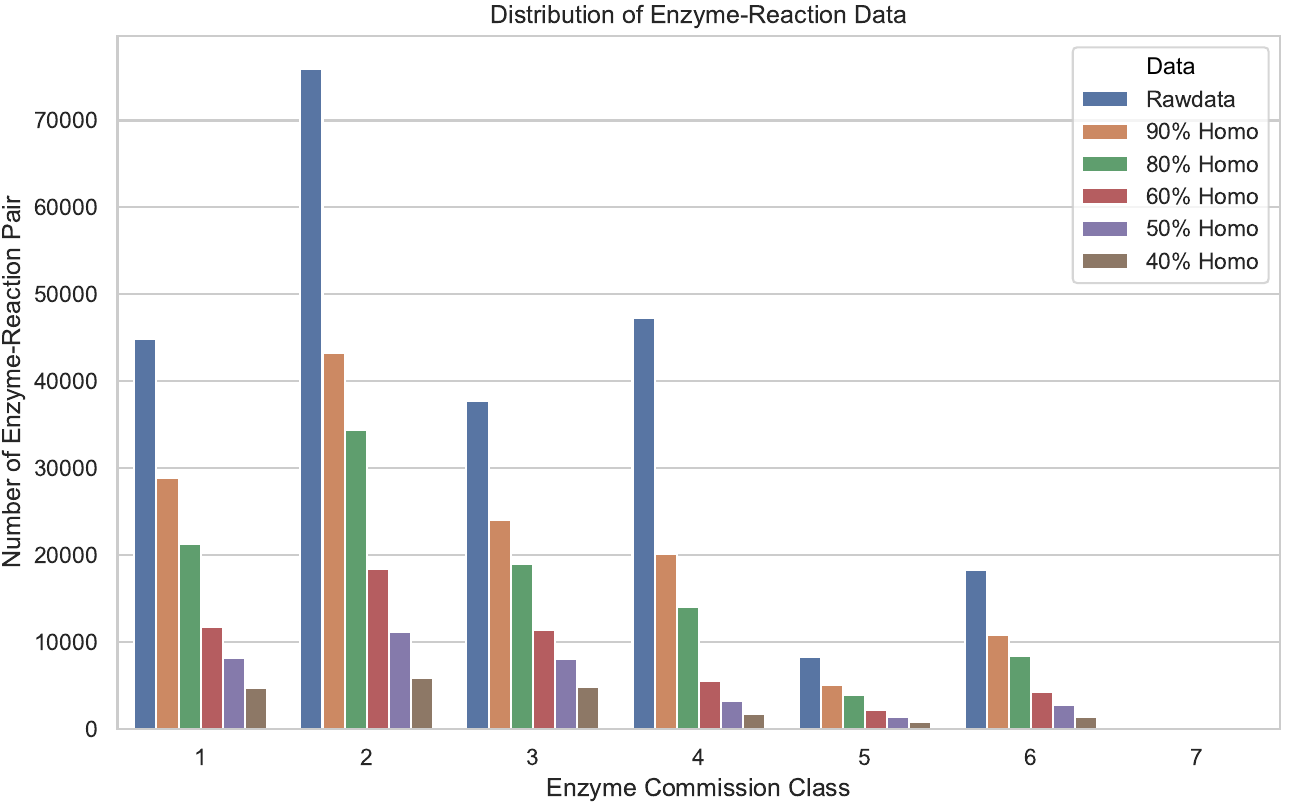}}
\vspace{-0.3cm}
{
  \caption{Distribution of enzyme-reaction pairs over EC-class.}
  \vspace{-0.5cm}
  \label{fig:data.stats}
}
\end{figure*}

\textbf{Debiasing.}
Please refer to Sec.~\ref{sec:enzymeflow.dataset}. We provide data statistics, including the EC-class distribution, in Table~\ref{tab:data.stats}, and visualize the distribution in Figure~\ref{fig:data.stats}.

From the data, we observe that EC1, EC2, EC3, and EC4 contribute the most enzyme-reaction pairs to our dataset. Specifically, EC1 refers to oxidation/reduction reactions, involving the transfer of hydrogen, oxygen atoms, or electrons from one substance to another. EC2 involves the transfer of a functional group (such as methyl, acyl, amino, or phosphate) from one substance to another. EC3 is associated with the formation of two products from a substrate through hydrolysis, while EC4 involves the non-hydrolytic addition or removal of groups from substrates, potentially cleaving C-C, C-N, C-O, or C-S bonds. 
Our dataset distribution closely follows the natural enzyme-reaction enzyme commission class distribution, with Transferases (EC2) being the most dominant.

\section{Work in Progress: Enzyme Pocket-Reaction Recruitment with Enzyme CLIP Model}
\label{app:enzyme.retrieval}

In addition to evaluating the catalytic pockets generated from the functional and structural perspectives, we may raise a key question of how we \textit{ quantitatively} determine whether the generated pockets can catalyze a specific reaction. To answer it, we are working to train an enzyme-reaction CLIP model using enzyme-reaction pairs (with pocket-specific information) from the $60\%$-clustered data, excluding the 100 evaluation samples from training. All enzymes not annotated to catalyze a specific reaction are treated as negative samples, following the approach in \cite{yang2024care, mikhael2024clipzyme}. For the $100$ generated catalytic pockets of each reaction, we select the \texttt{Top-1} pocket with the highest \texttt{TM-score} for evaluation using the enzyme CLIP model.


\begin{figure*}[ht!]
\centering
{
\includegraphics[width=0.98\textwidth]{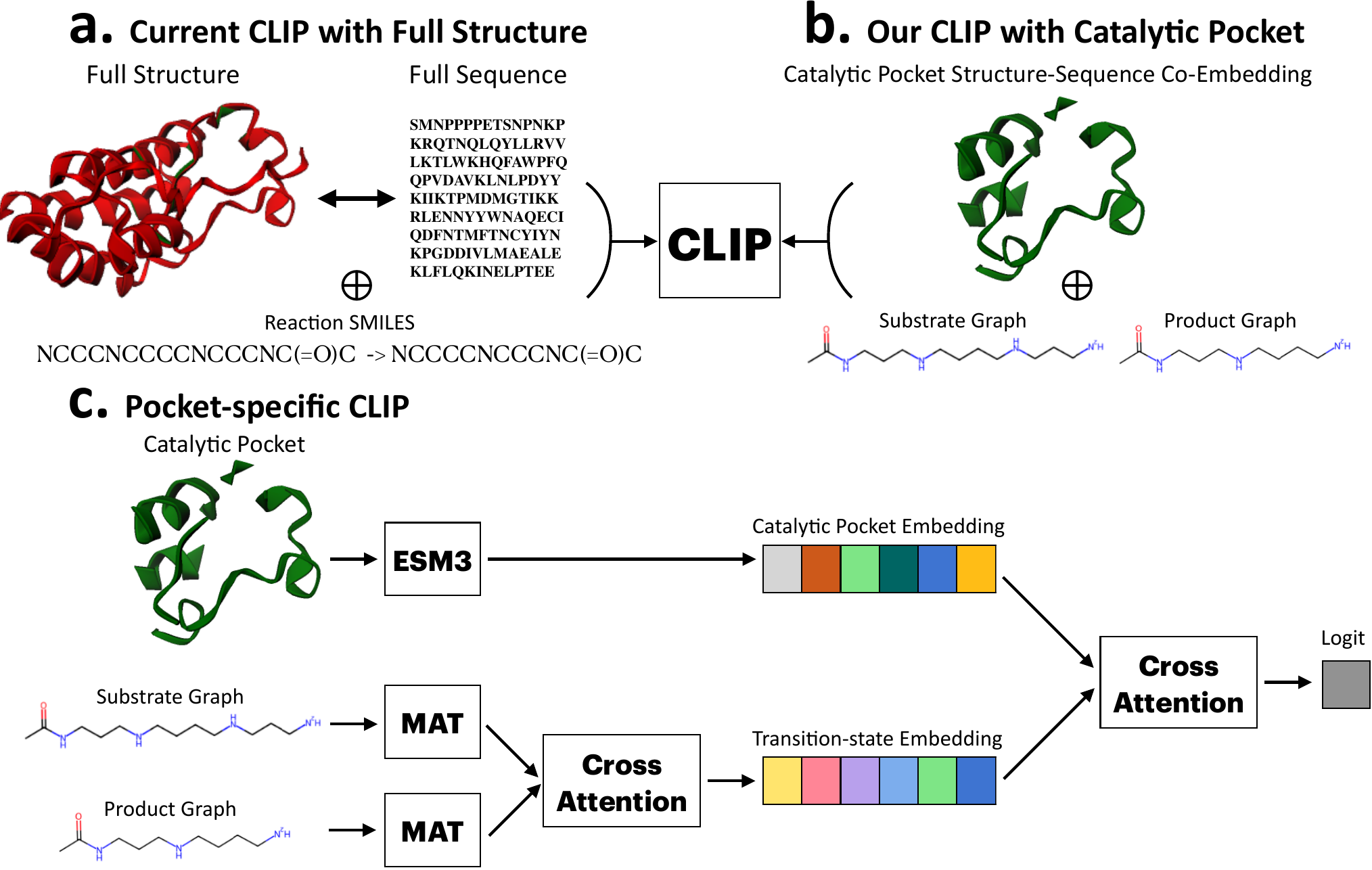}}
{
\vspace{-0.3cm}
  \caption{Enzyme-Reaction CLIP model comparison. (a) Existing CLIP models use the full enzyme structure or full enzyme sequence, paired with reaction SMILES as input. (b) Our pocket-specific CLIP model focuses on catalytic pockets, using both their structures and sequences paired with molecular graphs of reactions. The pocket-specific CLIP approach learns from enzyme active sites, which exhibit higher functional concentration. (c) Overview of Pocket-specific CLIP model.}
  \label{fig:clip_full}
}
\end{figure*}

\textbf{Pocket-specific CLIP.}
Unlike existing methods that typically train on full enzyme structures or sequences \citep{yu2023enzyme, mikhael2024clipzyme}, our pocket-specific CLIP approach is designed to focus specifically on catalytic pockets, including both their structures and sequences, paired with molecular graphs of catalytic reactions (illustrated in Fig.~\ref{fig:clip_full}). As shown in Fig.~\ref{fig:pocket.extraction}(b), catalytic pockets are usually the regions that exhibit high functional concentration, while the remaining parts tend to be less functionally important. Therefore, focusing on catalytic pockets is more applicable and effective for enzyme CLIP models. The advantage of the pocket-specific CLIP is that it learns from active sites that are highly meaningful both structurally and sequentially.

We illustrate our pocket-specific enzyme CLIP approach in Fig.~\ref{fig:clip_full}.
In our pocket-specific CLIP model, we encode the pocket structure and sequence using ESM3 \citep{hayes2024simulating}, and the substrate and product molecular graphs using MAT \citep{maziarka2020molecule}. Cross-attention is applied to compute the transition state of the reaction, capturing the transformation of the substrate into the product, as proposed in \cite{hua2024reactzyme}. This is followed by another cross-attention mechanism to learn the interactions between the catalytic pocket and the reaction. The model is trained by enforcing high logits for positive enzyme-reaction pairs and low logits for negative enzyme-reaction pairs.

\textbf{Metrics.}
To evaluate the catalytic ability of the designed pockets for a given reaction, we employ retrieval-based ranking as proposed in \cite{hua2024reactzyme}. This ranking-based evaluation ensures fairness and minimizes biases. The metrics include: \texttt{Top-k Acc}, which quantifies the proportion of instances in which the catalytic pocket is ranked within the CLIP’s top-k predictions; \texttt{Mean Rank},  which calculates the average position of the pocket in the retrieval list; Mean Reciprocal Rank (\texttt{MRR}), which measures how quickly the pocket is retrieved by averaging the reciprocal ranks of the first correct pocket across all reactions. These metrics help assess whether a catalytic pocket designed for a specific reaction ranks highly in the recruitment list, indicating its potential to catalyze the reaction.

\subsection{Inpainting Catalytic Pocket with ESM3 for Full Enzyme Recruitment}

\begin{figure*}[ht!]
\centering
{
\includegraphics[width=0.98\textwidth]{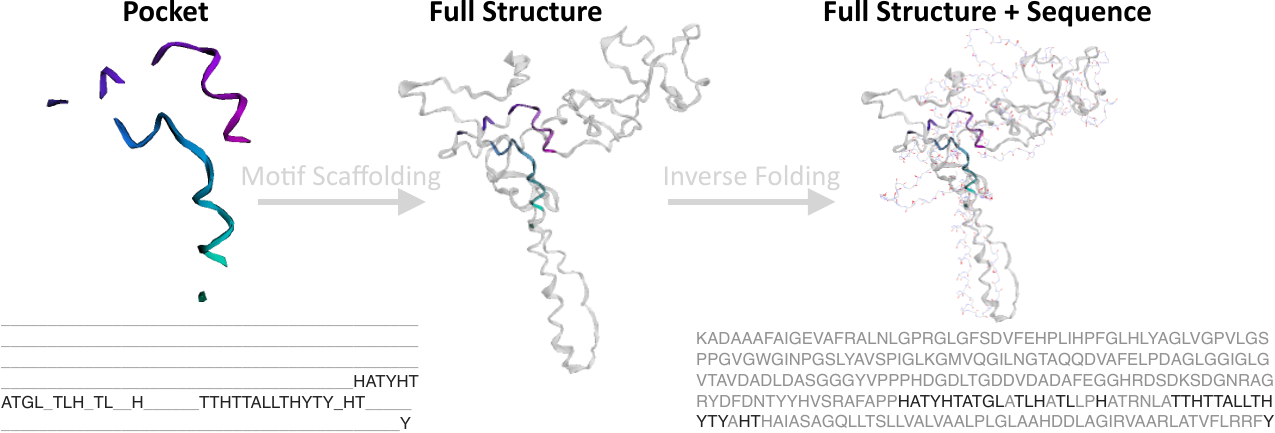}}
{
\vspace{-0.3cm}
  \caption{Inpainting catalytic pocket using ESM3. }
  \label{fig:esm3.impainting}
}
\end{figure*}

ESM3 \citep{hayes2024simulating} can inpaint missing structures and sequences with functional motifs. In this context, we train a separate full enzyme CLIP model for the enzyme recruitment task. This model is trained using the same $60\%$-clustered data but incorporates full enzyme structures and sequences. For generated catalytic pockets and those in the evaluation set, we use ESM3 to inpaint them, completing the structures and sequences predicted by ESM3. These ESM3-inpainted enzymes are then evaluated using the full enzyme CLIP model, applying the same retrieval-based ranking metrics as before. We illustrate the catalytic pocket inpainting pipeline in Fig.~\ref{fig:esm3.impainting}.

In conclusion, we are developing a pocket-specific enzyme CLIP model for pocket-based enzyme recruitment tasks and a full-enzyme CLIP model using ESM3 for inpainting and pocket scaffolding in full enzyme recruitment tasks. However, we recognize that directly using ESM3 for catalytic pocket inpainting lacks domain-specific knowledge, making fine-tuning necessary. To address this, we are working on a fine-tuning open-source large biological model, \textit{e.g.,} Genie2 \citep{lin2024out}, on our EnzymeFill dataset. Genie2, pre-trained on FoldSeek-clustered AlphaFold- and Protein-DataBank proteins for \textit{de novo} protein design and (multi-)motif scaffolding, aligns well with our catalytic pocket scaffolding task. Fine-tuning Genie2 on EnzymeFill will enhance its performance in catalytic pocket inpainting. The development of EnzymeFlow, aimed at achieving an AI-driven automated enzyme design platform, is discussed in App.~\ref{app:future.work}.

\begin{figure*}[ht!]
\centering
{
\includegraphics[width=1\textwidth]{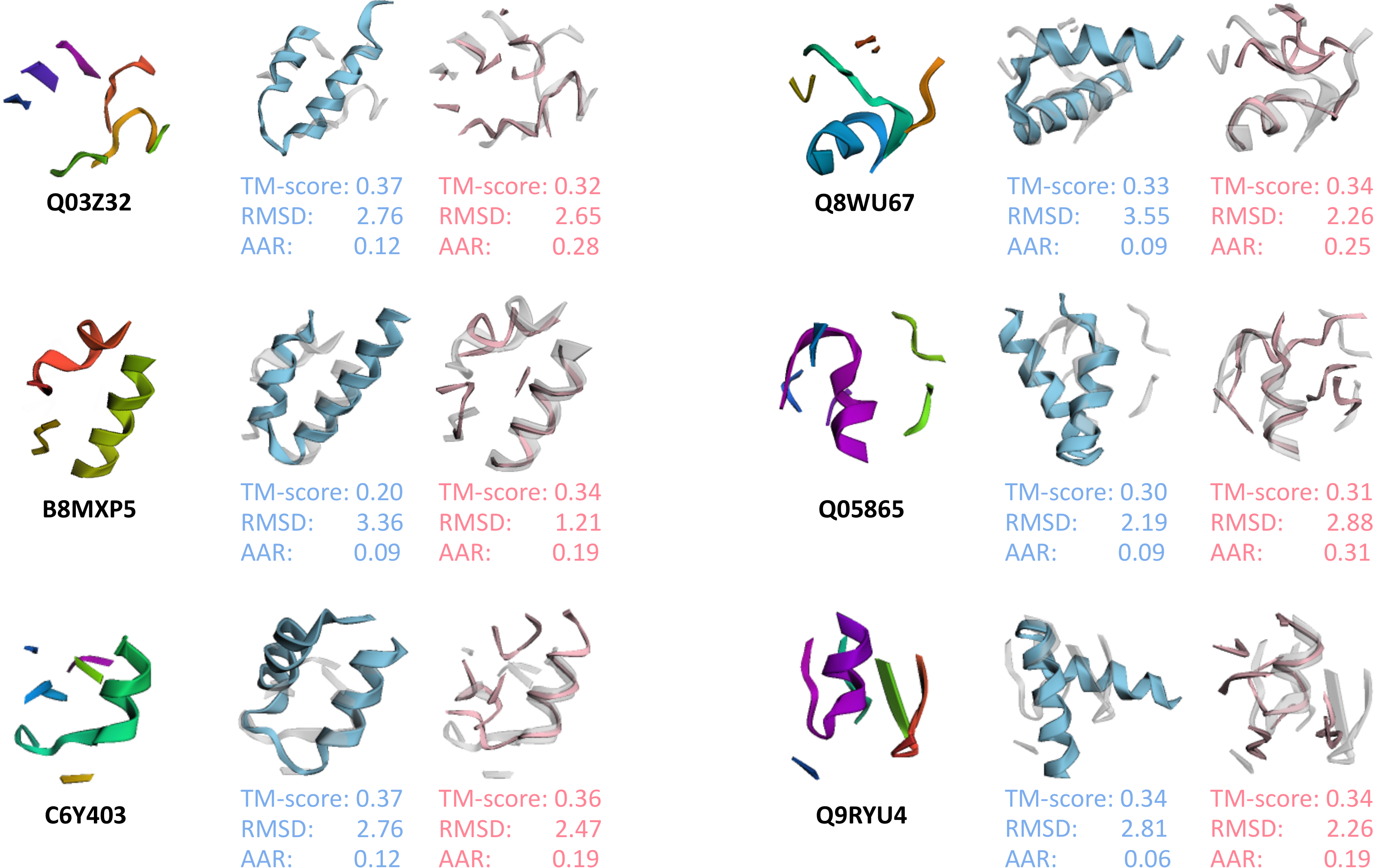}}
\vspace{-0.3cm}
{
  \caption{Visualization and comparison between RFDiffusionAA-designed pockets and EnzymeFlow-designed pockets after superimposition with ground-truth pockets. \textcolor{Lavender}{Light color represents EnzymeFlow-designed pockets}, \textcolor{SkyBlue}{blue color represents RFDiffusionAA-designed pockets}, \textcolor{LimeGreen}{spectral color represents the ground-truth reference pockets}. \texttt{TM-score}, \texttt{RMSD}, 
  \texttt{AAR} are reported.}
  \label{fig:rfdiff.pocket}
}
\end{figure*}

\section{RFDiffusionAA-design vs. EnzymeFlow-design}
\label{app:RFDiffusionAA.visual}
\vspace{-0.2cm}
In Fig.~\ref{fig:rfdiff.pocket}, we visualize and compare the RFDiffusionAA-generated pockets \citep{krishna2024generalized} with EnzymeFlow-generated catalytic pockets, both aligned to the ground-truth reference pockets. In RFDiffusionAA, the generation is conditioned on the substrate conformation, and the pocket sequence is computed post hoc using LigandMPNN \citep{dauparas2023atomic}. In contrast, EnzymeFlow conditions the generation on the reaction, with the pocket sequence co-designed alongside the pocket structure. In addition to visualization, we report \texttt{TM-score}, \texttt{RMSD}, and \texttt{AAR}, where EnzymeFlow outperforms RFDiffusionAA across all three metrics, demonstrating EnzymeFlow's ability to generate more structurally valid enzyme catalytic pockets.

\section{EnzymeFlow Neural Network Implementation}
\vspace{-0.2cm}
The equivariant neural network is based on the Invariant Point Attention (IPA) implemented in AlphaFold2 \citep{jumper2021highly}. In the following, we detail how enzyme catalytic pockets, substrate molecules, product molecules, EC-class, and co-evolution interact within our network.

The code for EnzymeFlow main network follows directly:
\begin{lstlisting}[language=Python, caption=Pytorch Implementation of EnzymeFlow Main Network.]
import functools as fn
import math

import torch
import torch.nn as nn
from torch.nn import functional as F

from ofold.utils.rigid_utils import Rigid

from model import ipa_pytorch
from flowmatch.data import all_atom
from flowmatch.data import utils as du

## EnzymeFlow Main Network

## (8) Distogram
def calc_distogram(pos, min_bin, max_bin, num_bins):
    dists_2d = torch.linalg.norm(pos[:, :, None, :] - pos[:, None, :, :], axis=-1)[
        ..., None
    ]
    lower = torch.linspace(min_bin, max_bin, num_bins, device=pos.device)
    upper = torch.cat([lower[1:], lower.new_tensor([1e8])], dim=-1)
    dgram = ((dists_2d > lower) * (dists_2d < upper)).type(pos.dtype)
    return dgram


## (7) Index Embedding
def get_index_embedding(indices, embed_size, max_len=2056):
    K = torch.arange(embed_size // 2, device=indices.device)
    pos_embedding_sin = torch.sin(
        indices[..., None] * math.pi / (max_len ** (2 * K[None] / embed_size))
    ).to(indices.device)
    pos_embedding_cos = torch.cos(
        indices[..., None] * math.pi / (max_len ** (2 * K[None] / embed_size))
    ).to(indices.device)
    pos_embedding = torch.cat([pos_embedding_sin, pos_embedding_cos], axis=-1)
    return pos_embedding


## (6) Time Embedding
def get_timestep_embedding(timesteps, embedding_dim, max_positions=10000):
    assert len(timesteps.shape) == 1
    timesteps = timesteps * max_positions
    half_dim = embedding_dim // 2
    emb = math.log(max_positions) / (half_dim - 1)
    emb = torch.exp(
        torch.arange(half_dim, dtype=torch.float32, device=timesteps.device) * -emb
    )
    emb = timesteps.float()[:, None] * emb[None, :]
    emb = torch.cat([torch.sin(emb), torch.cos(emb)], dim=1)
    if embedding_dim % 2 == 1:  # zero pad
        emb = F.pad(emb, (0, 1), mode="constant")
    assert emb.shape == (timesteps.shape[0], embedding_dim)
    return emb


## (5) Edge Feature Network
class EdgeFeatureNet(nn.Module):
    def __init__(self, module_cfg):
        super(EdgeFeatureNet, self).__init__()
        self._cfg = module_cfg

        self.c_s = self._cfg.embed.c_s
        self.c_z = self._cfg.embed.c_z
        self.feat_dim = self._cfg.embed.feat_dim

        self.linear_s_p = nn.Linear(self.c_s, self.feat_dim)
        self.linear_relpos = nn.Linear(self.feat_dim, self.feat_dim)

        total_edge_feats = self.feat_dim * 3 + self._cfg.embed.num_bins * 2 + 2

        self.edge_embedder = nn.Sequential(
            nn.Linear(total_edge_feats, self.c_z),
            nn.ReLU(),
            nn.Linear(self.c_z, self.c_z),
            nn.ReLU(),
            nn.Linear(self.c_z, self.c_z),
            nn.LayerNorm(self.c_z),
        )

    def embed_relpos(self, r):
        d = r[:, :, None] - r[:, None, :]
        pos_emb = get_index_embedding(d, self.feat_dim, max_len=2056)
        return self.linear_relpos(pos_emb)

    def _cross_concat(self, feats_1d, num_batch, num_res):
        return torch.cat([
            torch.tile(feats_1d[:, :, None, :], (1, 1, num_res, 1)),
            torch.tile(feats_1d[:, None, :, :], (1, num_res, 1, 1)),
        ], dim=-1).float().reshape([num_batch, num_res, num_res, -1])

    def forward(self, s, t, sc_t, edge_mask, flow_mask):
        # Input: [b, n_res, c_s]
        num_batch, num_res, _ = s.shape

        # [b, n_res, c_z]
        p_i = self.linear_s_p(s)
        cross_node_feats = self._cross_concat(p_i, num_batch, num_res)

        # [b, n_res]
        r = torch.arange(
            num_res, device=s.device).unsqueeze(0).repeat(num_batch, 1)
        relpos_feats = self.embed_relpos(r)

        dist_feats = calc_distogram(
            t, min_bin=1e-3, max_bin=20.0, num_bins=self._cfg.embed.num_bins)
        sc_feats = calc_distogram(
            sc_t, min_bin=1e-3, max_bin=20.0, num_bins=self._cfg.embed.num_bins)

        all_edge_feats = [cross_node_feats, relpos_feats, dist_feats, sc_feats]

        diff_feat = self._cross_concat(flow_mask[..., None], num_batch, num_res)
        all_edge_feats.append(diff_feat)
        
        edge_feats = self.edge_embedder(torch.concat(all_edge_feats, dim=-1).to(torch.float))
        edge_feats *= edge_mask.unsqueeze(-1)
        return edge_feats
    

## (4) Node Feature Network
class NodeFeatureNet(nn.Module):
    def __init__(self, module_cfg):
        super(NodeFeatureNet, self).__init__()
        self._cfg = module_cfg
        self.c_s = self._cfg.embed.c_s
        self.c_pos_emb = self._cfg.embed.c_pos_emb
        self.c_timestep_emb = self._cfg.embed.c_timestep_emb
        embed_size = self.c_pos_emb + self.c_timestep_emb * 2 + 1

        self.aatype_embedding = nn.Embedding(21, self.c_s) # Always 21 because of 20 amino acids + 1 for unk
        embed_size += self.c_s + self.c_timestep_emb + self._cfg.num_aa_type
            
        self.linear = nn.Sequential(
            nn.Linear(embed_size, self.c_s),
            nn.ReLU(),
            nn.Linear(self.c_s, self.c_s),
            nn.ReLU(),
            nn.Linear(self.c_s, self.c_s),
            nn.LayerNorm(self.c_s),
        )

    def embed_t(self, timesteps, mask):
        timestep_emb = get_timestep_embedding(
            timesteps,
            self.c_timestep_emb,
            max_positions=2056
        )[:, None, :].repeat(1, mask.shape[1], 1)
        return timestep_emb * mask.unsqueeze(-1)

    def forward(
            self,
            *,
            t,
            res_mask,
            flow_mask,
            pos,
            aatypes,
            aatypes_sc,
        ):
        # [b, n_res, c_pos_emb]
        pos_emb = get_index_embedding(pos, self.c_pos_emb, max_len=2056)
        pos_emb = pos_emb * res_mask.unsqueeze(-1)

        # [b, n_res, c_timestep_emb]
        input_feats = [
            pos_emb,
            flow_mask[..., None],
            self.embed_t(t, res_mask),
            self.embed_t(t, res_mask)
        ]
        input_feats.append(self.aatype_embedding(aatypes))
        input_feats.append(self.embed_t(t, res_mask))
        input_feats.append(aatypes_sc)
        return self.linear(torch.cat(input_feats, dim=-1))


## (3) Distance Embedder
class DistEmbedder(nn.Module):
    def __init__(self, model_conf):
        super(DistEmbedder, self).__init__()
        torch.set_default_dtype(torch.float32)
        self._model_conf = model_conf
        self._embed_conf = model_conf.embed

        edge_embed_size = self._model_conf.edge_embed_size

        self.dist_min = self._model_conf.bb_ligand_rbf_d_min
        self.dist_max = self._model_conf.bb_ligand_rbf_d_max
        self.num_rbf_size = self._model_conf.num_rbf_size
        self.edge_embedder = nn.Sequential(
            nn.Linear(self.num_rbf_size, edge_embed_size),
            nn.ReLU(),
            nn.Linear(edge_embed_size, edge_embed_size),
            nn.ReLU(),
            nn.Linear(edge_embed_size, edge_embed_size),
            nn.LayerNorm(edge_embed_size),
        )

        mu = torch.linspace(self.dist_min, self.dist_max, self.num_rbf_size)
        self.mu = mu.reshape([1, 1, 1, -1])
        self.sigma = (self.dist_max - self.dist_min) / self.num_rbf_size

    def coord2dist(self, coord, edge_mask):
        n_batch, n_atom = coord.size(0), coord.size(1)
        radial = torch.sum((coord.unsqueeze(1) - coord.unsqueeze(2)) ** 2, dim=-1)
        dist = torch.sqrt(
                radial + 1e-10
            ) * edge_mask

        radial = radial * edge_mask
        return radial, dist
    
    def rbf(self, dist):
        dist_expand = torch.unsqueeze(dist, -1)
        _mu = self.mu.to(dist.device)
        rbf = torch.exp(-(((dist_expand - _mu) / self.sigma) ** 2))
        return rbf

    def forward(
        self,
        rigid,
        ligand_pos,
        bb_ligand_mask,
    ):
        curr_bb_pos = all_atom.to_atom37(Rigid.from_tensor_7(torch.clone(rigid)))[-1][:, :, 1].to(ligand_pos.device)

        curr_bb_lig_pos = torch.cat([curr_bb_pos, ligand_pos], dim=1)
        edge_mask = bb_ligand_mask.unsqueeze(dim=1) * bb_ligand_mask.unsqueeze(dim=2)
        
        radial, dist = self.coord2dist(
                            coord=curr_bb_lig_pos, 
                            edge_mask=edge_mask,
                        )


        edge_embed = self.rbf(dist) * edge_mask[..., None]
        edge_embed = self.edge_embedder(edge_embed.to(torch.float))

        return edge_embed

        
## (2) Cross-Attentiom
class CrossAttention(nn.Module):
    def __init__(self, query_input_dim, key_input_dim, output_dim):
        super(CrossAttention, self).__init__()
        self.out_dim = output_dim
        self.W_Q = nn.Linear(query_input_dim, output_dim)
        self.W_K = nn.Linear(key_input_dim, output_dim)
        self.W_V = nn.Linear(key_input_dim, output_dim)
        self.scale_val = self.out_dim ** 0.5
        self.softmax = nn.Softmax(dim=-1)
    
    def forward(self, query_input, key_input, value_input, query_input_mask=None, key_input_mask=None):
        query = self.W_Q(query_input)
        key = self.W_K(key_input)
        value = self.W_V(value_input)

        attn_weights = torch.matmul(query, key.transpose(1, 2)) / self.scale_val
        attn_mask = query_input_mask.unsqueeze(-1) * key_input_mask.unsqueeze(-1).transpose(1, 2)
        attn_weights = attn_weights.masked_fill(attn_mask == False, -1e9)
        attn_weights = self.softmax(attn_weights)
        output = torch.matmul(attn_weights, value)
        
        return output, attn_weights


## (1) Protein-Ligand Network
class ProteinLigandNetwork(nn.Module):
    def __init__(self, model_conf):
        super(ProteinLigandNetwork, self).__init__()
        torch.set_default_dtype(torch.float32)
        self._model_conf = model_conf

        # Input Node Embedder
        self.node_feature_net = NodeFeatureNet(model_conf)

        # Input Edge Embedder
        self.edge_feature_net = EdgeFeatureNet(model_conf)

        # 3D Molecule GNN
        self.mol_embedding_layer = MolEmbedder(model_conf)

        # Invariant Point Attention (IPA) Network
        self.ipanet = ipa_pytorch.IpaNetwork(model_conf)

        # Node Fusion
        self.node_embed_size = self._model_conf.node_embed_size
        self.node_embedder = nn.Sequential(
            nn.Embedding(self._model_conf.num_aa_type, self.node_embed_size),
            nn.ReLU(),
            nn.Linear(self.node_embed_size, self.node_embed_size),
            nn.LayerNorm(self.node_embed_size),
        )
        self.node_fusion = nn.Sequential(
            nn.Linear(self.node_embed_size + self.node_embed_size, self.node_embed_size),
            nn.ReLU(),
            nn.Linear(self.node_embed_size, self.node_embed_size),
            nn.LayerNorm(self.node_embed_size),
        )

        # Backbone-Substrate Fusion
        self.bb_lig_fusion = CrossAttention(
                query_input_dim=self.node_embed_size,
                key_input_dim=self.node_embed_size,
                output_dim=self.node_embed_size,
        )

        # Edge Fusion
        self.edge_embed_size = self._model_conf.edge_embed_size
        self.edge_dist_embedder = DistEmbedder(model_conf)

        # Amino Acid Prediction Network
        self.aatype_pred_net = nn.Sequential(
                nn.Linear(self.node_embed_size, self.node_embed_size),
                nn.ReLU(),
                nn.Linear(self.node_embed_size, self.node_embed_size),
                nn.ReLU(),
                nn.Linear(self.node_embed_size, model_conf.num_aa_type), 
        )

        if self._model_conf.flow_msa:
            # Co-Evolution Embedder
            self.msa_embedding_layer = CoEvoFormer(model_conf)

            # Coevo-Backbone-Substrate Fusion
            self.msa_bb_lig_fusion = CrossAttention(
                query_input_dim=model_conf.msa.msa_embed_size,
                key_input_dim=self.node_embed_size,
                output_dim=self.node_embed_size,
            )

            # Coevo Prediction Network
            self.msa_pred = nn.Sequential(
                nn.Linear(self.node_embed_size, self.node_embed_size),
                nn.SiLU(),
                nn.Linear(self.node_embed_size, self.node_embed_size),
                nn.SiLU(),
                nn.Linear(self.node_embed_size, model_conf.msa.num_msa_vocab),
            )

        if self._model_conf.ec:
            # EC Embedder
            self.ec_embedding_layer = nn.Sequential(
                nn.Embedding(model_conf.ec.num_ec_class, model_conf.ec.ec_embed_size),
                nn.SiLU(),
                nn.Linear(model_conf.ec.ec_embed_size, model_conf.ec.ec_embed_size),
                nn.LayerNorm(model_conf.ec.ec_embed_size),
            )

            # EC-Backbone-Substrate Fusion
            self.ec_bb_lig_fusion = CrossAttention(
                query_input_dim=model_conf.ec.ec_embed_size,
                key_input_dim=self.node_embed_size,
                output_dim=self.node_embed_size,
            )

            # EC Prediction Network
            self.ec_pred = nn.Sequential(
                nn.Linear(self.node_embed_size, self.node_embed_size),
                nn.SiLU(),
                nn.Linear(self.node_embed_size, self.node_embed_size),
                nn.SiLU(),
                nn.Linear(self.node_embed_size, model_conf.ec.num_ec_class),
            )

        self.condition_generation = self._model_conf.guide_by_condition
        if self.condition_generation:
            # 2D Molecule GNN
            self.guide_ligand_mpnn = MolEmbedder2D(model_conf)

            # Backbone-Product Fusion
            self.guide_bb_lig_fusion = CrossAttention(
                query_input_dim=self.node_embed_size,
                key_input_dim=self.node_embed_size,
                output_dim=self.node_embed_size,
            )

    def forward(self, input_feats, use_context=False):
        # Frames as [batch, res, 7] tensors.
        bb_mask = input_feats["res_mask"].type(torch.float32)  # [B, N]
        flow_mask = input_feats["flow_mask"].type(torch.float32)
        edge_mask = bb_mask[..., None] * bb_mask[..., None, :]

        n_batch, n_res = bb_mask.shape

        # Encode Backbone Nodes with Input Node Embedder
        init_bb_node_embed = self.node_feature_net(
            t=input_feats["t"],
            res_mask=bb_mask,
            flow_mask=flow_mask,
            pos=input_feats["seq_idx"],
            aatypes=input_feats["aatype_t"],
            aatypes_sc=input_feats["sc_aa_t"],
        )

        # Encode Backbone Edges with Input Edge Embedder
        init_bb_edge_embed = self.edge_feature_net(
            s=init_bb_node_embed,
            t=input_feats["trans_t"],
            sc_t=input_feats["sc_ca_t"],
            edge_mask=edge_mask,
            flow_mask=flow_mask,
        )

        # Masking Padded Residues
        bb_node_embed = init_bb_node_embed * bb_mask[..., None]
        bb_edge_embed = init_bb_edge_embed * edge_mask[..., None]

        # AminoAcid embedding
        bb_aa_embed = self.node_embedder(input_feats["aatype_t"]) * bb_mask[..., None]
        bb_aa_embed = torch.cat([bb_aa_embed, bb_node_embed], dim=-1)
        # Backbone-AminoAcid Fusion
        bb_node_embed = self.node_fusion(bb_aa_embed)
        bb_node_embed = bb_node_embed * bb_mask[..., None]
        
        # Initialze Substrate Masking
        lig_mask = input_feats["ligand_mask"]
        lig_edge_mask = lig_mask[..., None] * lig_mask[..., None, :]
        # Encode Substrate with 3D Molecule GNN
        lig_init_node_embed, _ = self.mol_embedding_layer(
                ligand_atom=input_feats["ligand_atom"],
                ligand_pos=input_feats["ligand_pos"],
                edge_mask=lig_edge_mask,
            )
        lig_node_embed = lig_init_node_embed * lig_mask[..., None]

        # Backbone-Substrate Fusion
        bb_lig_rep, _ = self.bb_lig_fusion(
                                query_input=bb_node_embed, 
                                key_input=lig_node_embed, 
                                value_input=lig_node_embed, 
                                query_input_mask=bb_mask, 
                                key_input_mask=lig_mask,
                            )

        # Residue Connection
        bb_node_embed = bb_node_embed + bb_lig_rep

        # Conditioning on Product Molecule
        if self.condition_generation:
            # Encode Product with 2D Molecule GNN
            guide_ligand_rep = self.guide_ligand_mpnn(
                                mol_atom=input_feats["guide_ligand_atom"],
                                mol_edge=input_feats["guide_ligand_edge_index"],
                                mol_edge_feat=input_feats["guide_ligand_edge"],
                                mol_atom_mask=input_feats["guide_ligand_atom_mask"],
                                mol_edge_mask=input_feats["guide_ligand_edge_mask"],
                            ).unsqueeze(1)

            # Initialze Product Masking
            guide_ligand_mask = input_feats["guide_ligand_atom_mask"][:, 0:1]
            # Backbone-Product Fusion
            bb_guide_lig_rep, _ = self.guide_bb_lig_fusion(
                                    query_input=bb_node_embed, 
                                    key_input=guide_ligand_rep, 
                                    value_input=guide_ligand_rep, 
                                    query_input_mask=bb_mask, 
                                    key_input_mask=guide_ligand_mask,
                                )

            # Residue Connection
            bb_node_embed = bb_node_embed + bb_guide_lig_rep

        # Initialze Backbone-Substrate Masking
        bb_ligand_mask = torch.cat([bb_mask, lig_mask], dim=-1)
        # Backbone-Substrate Distance Embedding
        bb_lig_edge = self.edge_dist_embedder(
            rigid=input_feats["rigids_t"],
            ligand_pos=input_feats["ligand_pos"],
            bb_ligand_mask=bb_ligand_mask,
        )

        # Backbone-Backbone-Product Edge Fusion
        bb_edge_embed = bb_edge_embed + bb_lig_edge[:, :n_res, :n_res, :]

        # Masking Padded Residues
        bb_node_embed = bb_node_embed[:, :n_res, :] * bb_mask[..., None]
        bb_edge_embed = bb_edge_embed[:, :n_res, :n_res, :] * edge_mask[..., None]

        # Run IPA Network
        model_out = self.ipanet(bb_node_embed, bb_edge_embed, input_feats)
        node_embed = model_out["node_embed"] * bb_mask[..., None]

        # Amino Acid Prediction with Amino Acid Prediction Network
        aa_pred = self.aatype_pred_net(node_embed) * bb_mask[..., None]

        if self._model_conf.flow_msa:
            # Encode Coevo with Co-Evolution Embedder
            msa_mask = input_feats["msa_mask"]
            msa_embed = self.msa_embedding_layer(input_feats["msa_t"], msa_mask=msa_mask) * msa_mask[..., None] #[B, N_msa, N_token, D]
            msa_rep = msa_embed.sum(dim=1) / (msa_mask[..., None].sum(dim=1) + 1e-10) #[B, 1, D]
            _msa_mask = msa_mask[:, 0] #torch.ones_like(msa_rep[..., 0]).to(msa_embed.device)

            # Coevo-Backbone Fusion
            msa_rep, _ = self.msa_bb_lig_fusion(
                                        query_input=msa_rep, 
                                        key_input=node_embed, 
                                        value_input=node_embed, 
                                        query_input_mask=_msa_mask, 
                                        key_input_mask=bb_mask,
                                    )

            # Coevo Prediction with Coevo Prediction Network
            msa_pred = self.msa_pred(msa_rep)
        
        if self._model_conf.flow_ec:
            # Encode EC with EC Embedder
            ec_embed = self.ec_embedding_layer(input_feats["ec_t"])
            ec_mask = torch.ones_like(ec_embed[..., 0]).to(ec_embed.device)

            # EC-Backbone Fusion
            ec_rep, _ = self.ec_bb_lig_fusion(
                                    query_input=ec_embed, 
                                    key_input=node_embed, 
                                    value_input=node_embed, 
                                    query_input_mask=ec_mask, 
                                    key_input_mask=bb_mask,
                                )

            # EC Prediction with EC Prediction Network
            ec_rep = ec_rep.reshape(n_batch, -1)
            ec_pred = self.ec_pred(ec_rep)

        # Main Network Ouput
        pred_out = {
            "amino_acid": aa_pred,
            "rigids_tensor": model_out["rigids"],
        }
        
        if self._model_conf.flow_msa:
            pred_out["msa"] = msa_pred * _msa_mask[..., None]

        if self._model_conf.flow_ec:
            pred_out["ec"] = ec_pred

        pred_out["rigids"] = model_out["rigids"].to_tensor_7()
        return pred_out
\end{lstlisting}

\textit{Fun Fact:} While implementing enzyme-substrate and enzyme-product interactions by cross-attention fusion networks, we experimented with using PairFormer (with only 3-4 layers) as implemented in AlphaFold3 \citep{abramson2024accurate}. However, the computational load was immense—it would take years to run on our A40 GPU. Our fusion network turns to be a more efficient approach. It makes me wonder who has the resources to re-train AlphaFold3, given the heavy computational demands!

\end{document}